\pdfoutput=1

\documentclass[11pt]{article}

\usepackage[final]{acl}
\usepackage{times}
\usepackage{latexsym}

\usepackage[T1]{fontenc}

\usepackage[utf8]{inputenc}

\usepackage{microtype}

\usepackage{inconsolata}
\usepackage{changepage} 
\usepackage{graphicx}
\usepackage{xspace}
\usepackage{colortbl}
\usepackage{booktabs}
\usepackage{enumitem}
\usepackage{amssymb}
\usepackage{xcolor}
\usepackage{amsmath}
\usepackage{subfigure}
\usepackage{multirow}
\usepackage{listings}
\usepackage[normalem]{ulem}

\usepackage{tcolorbox}
\tcbuselibrary{listings,breakable,skins} 

\newtcblisting{promptbox}[2][]{
    title = {#2},               
    colback = gray!5!white,  
    colframe = gray!50!black,  
    coltitle = black,         
    colbacktitle=gray!30!white,  
    fonttitle = \bfseries\small,   
    fontupper = \ttfamily\small,     
    listing only,                  
    breakable,                   
    attach title to upper = \quad, 
    #1                             
}
\usepackage{cuted}
\usepackage{amssymb}

\definecolor{mygray}{RGB}{226, 226, 226}
\definecolor{myred}{RGB}{252, 142, 142}
\definecolor{mygreen}{RGB}{147, 255, 143}
\definecolor{myblue}{RGB}{144, 155, 255}
\definecolor{myyellow}{RGB}{253, 253, 143}
\definecolor{mypurple}{RGB}{255, 142, 250}

\title{Curse of Knowledge: When Complex Evaluation Context Benefits yet Biases LLM Judges}

\author{
 \textbf{Weiyuan Li\textsuperscript{1,2$\heartsuit$}},
 \textbf{Xintao Wang\textsuperscript{2,3$\dagger$}},
 \textbf{Siyu Yuan\textsuperscript{1,2}},
 \textbf{Rui Xu\textsuperscript{2,3}},
 \\
 \textbf{Jiangjie Chen\textsuperscript{4}},
 \textbf{Qingqing Dong\textsuperscript{5}},
\textbf{Yanghua Xiao\textsuperscript{2,3}},
\textbf{Deqing Yang\textsuperscript{1,2$\spadesuit$*}}
 \\
 \textsuperscript{1}School of Data Science, Fudan University,  
 \textsuperscript{2}Shanghai Key Laboratory of Data Science\\
 \textsuperscript{3}College of Computer Science and Artificial Intelligence, Fudan University \\
 \textsuperscript{4}ByteDance Seed,
 \textsuperscript{5}College of Cryptology and Cyber Science, Nankai University \\
 \textsuperscript{$\heartsuit$}\texttt{weiyuanli25@m.fudan.edu.cn}, 
\textsuperscript{$\spadesuit$}\texttt{yangdeqing@fudan.edu.cn}
 \\
}

\begin{document}
\maketitle
\newcommand{\method}{\textsc{ComplexEval}\xspace}
\renewcommand{\thefootnote}{\fnsymbol{footnote}}

\footnotetext[0]{$\dagger$  Project leader.}
\footnotetext[0]{*  Corresponding author.}
\renewcommand{\thefootnote}{\arabic{footnote}}

\begin{abstract}
As large language models (LLMs) grow more capable, they face increasingly diverse and complex tasks, making reliable evaluation challenging. The paradigm of LLMs as judges has emerged as a scalable solution, yet prior work primarily focuses on simple settings. Their reliability in complex tasks—where multi-faceted rubrics, unstructured reference answers, and nuanced criteria are critical—remains understudied. In this paper,  we constructed \textbf{\textsc{ComplexEval}}, a challenge benchmark designed to systematically expose and quantify \textit{Auxiliary Information Induced Biases.}  We systematically investigated and validated 5 previously unexplored biases across 12 basic and 3 advanced scenarios. Key findings reveal: (1) all evaluated models exhibit significant susceptibility to these biases, with bias magnitude scaling with task complexity; (2) notably, Large Reasoning Models (LRMs) show paradoxical vulnerability. Our in-depth analysis offers crucial insights for improving the accuracy and verifiability of evaluation signals, paving the way for more general and robust evaluation models.

\end{abstract}

\section{Introduction}

As Large Language Models (LLMs) face increasingly complex and diverse tasks~\citep{LLM:1}, the evaluation of these tasks becomes correspondingly more challenging (e.g., mathematical proofs, creative writing). The LLM-as-a-judge paradigm offers a low-cost and highly automated solution~\citep{judge-survey:1,judge-survey:2}, demonstrating strong alignment with human experts~\citep{judge:1}. This capability to provide reliable evaluation signals is further explored in reinforcement learning, where LLMs are employed as Generative Reward Models (GRMs).


\begin{figure}[t]
  \includegraphics[width=\columnwidth]{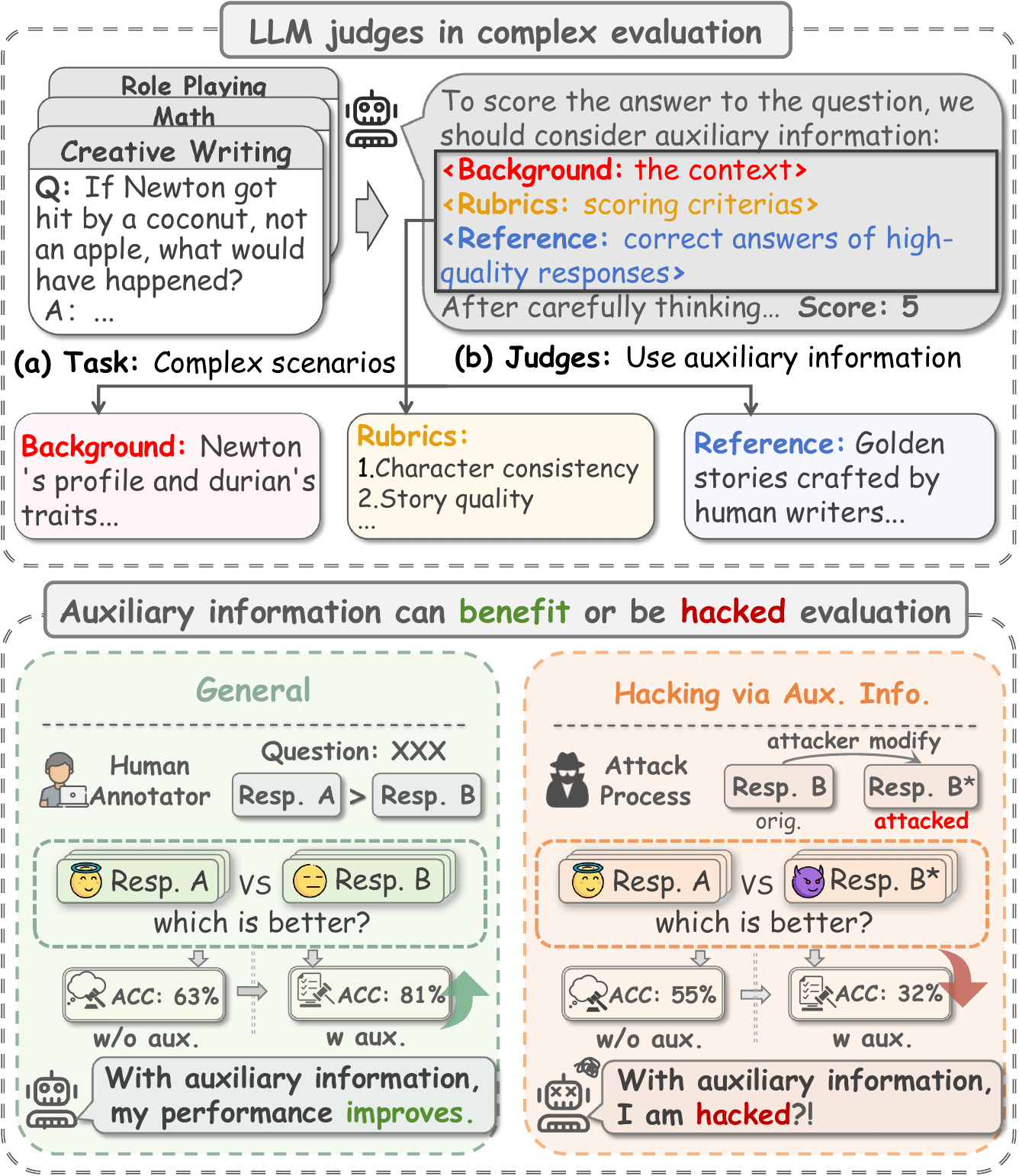}
  \caption{"When Help Become Harm": The figure shows that auxiliary information improves LLM judges but introduces new biases in complex evaluation}
  \label{fig:experiments_1}
\end{figure}

While LLM judges have the potential to address the growing demand for reliable and verifiable evaluation signals in complex tasks~\citep{Rubrics-RL-2,chen2024personapersonalizationsurveyroleplaying}, existing methods often fall short. Evaluating complex tasks often requires more guidance and provided knowledge~\citep{Reas-GRM:1,Complex-Judge:1}. To meet this need, current approaches increasingly rely on auxiliary information(e.g., reference answers~\citep{Reference:1}, rubrics~\citep{judge-rules:1,xiao2025humanizingmachinesrethinkingllm,huang2024on}), which can be external or self-generated. The use of auxiliary information enhances evaluation reliability and verifiability, however, it also introduces new challenges: LLM judges often struggles to follow complex instructions~\citep{RewardAnything} and can be influenced by specific patterns~\citep{OneToken}.Previous studies have identified various evaluation biases~\citep{Bias-Position:1,Bias-Length:1}, but they have generally failed to address these complex and information-rich evaluation tasks. We therefore investigate a crucial yet unexplored question: 


\noindent

\begin{adjustwidth}{0.4cm}{0.4cm}
\vspace{0.5\baselineskip}
\textit{Does the use of auxiliary information introduce bias in complex evaluation?}
\vspace{0.5\baselineskip}
\end{adjustwidth}

 

In this paper,  we study bias in complex evaluation, where reference,rubrics and background knowledge fundamentally shape judgment quality.
We propose \textsc{ComplexEval}, an adversarial framework designed to diagnose auxiliary information induced biases. The framework's two-tiered structure exposes these biases at varying levels of granularity: \textit{1)} \textbf{ComplexEval-Basic} conducts free-form attacks, broadly exploring bias induced by auxiliary information across twelve general tasks, 
while \textit{2)} \textbf{ComplexEval-Advanced} conducts fixed-pattern attacks, deeply analyzing specific bias types to study how rubrics, references and other auxiliary information corrupt evaluation mechanics based on three complex evaluation tasks. 



Our analysis reveals three key insights about auxiliary information induced bias: \textit{1)} All models exhibit significant biases when incorporating rubrics and reference answers, with severity scaling with task complexity; 
\textit{2)} We identify five specific bias patterns, including \textit{format bias}, \textit{solution fixation bias}, \textit{stereotype amplification bias}, \textit{criteria loophole bias} and \textit{criteria entanglement bias};
\textit{3)} While advanced reasoning models usually achieve better scores, they show greater vulnerability to these auxiliary information induced bias.

Our key contributions are as follows:
\begin{itemize} [leftmargin=*]
\item We propose to study LLM
judges in complex evaluation with rich auxiliary information, a widely applied but underexplored area.

\item We introduce ComplexEval (section~\ref{4}), the first systematic framework for studying how auxiliary information induced biases in complex evaluation of LLM judges .

\item Our analysis reveals that all LLM judges are heavily influenced by auxiliary information induced bias, with reasoning models being more vulnerable, and uncovers five new bias unique to complex evaluation (section~\ref{5}).

\end{itemize}

\section{Related work}

\subsection{LLM-as-a-Judge and GenRM}

LLMs have demonstrated remarkable capabilities in general domains~\citep{LLM:2}, prompting early research to explore their use for universal and scalable evaluation~\citep{Judge-Old-Work:1,Judge-Old-Work:2,Judge-Old-Work:3}--leading to the LLM-as-a-judge paradigm.  Subsequent empirical studies demonstrated that LLM-based evaluations align closely with expert judgments~\citep{judge:1,judge:2} and can even surpass human evaluators in fairness~\citep{judge:4}. Today, this approach has become mainstream, achieving robust performance across diverse domains~\citep{judge:5,judge:6} through structured evaluation methods~\citep{huang2025competing,li2025big5chatshapingllmpersonalities}.

Meanwhile, advances in RLHF highlight the demand for more general reward signals~\citep{Deepseek-math, Reas-GRM:1}. Unlike traditional scalar reward models, \citet{GenRM:1,GenRM:3} proposed GenRM, which bridges reward modeling and LLM-as-a-judge by leveraging LLM-generated rewards and iterative training. This approach demonstrates LLMs' potential to provide superior evaluation signals~\citep{GenRM:4}, particularly in complex scenarios.

\subsection{Enhancing LLM Judges Assessment in Complex Scenarios}

To improve LLM-as-a-judge evaluation in complex scenarios, common strategies include: \textbf{(1) auxiliary information enhancement}, using reference answers~\citep{Reference:1,Reference:2} or fine-grained rules~\citep{judge-rules:1,judge-rules:2,judge-rules:3}, and \textbf{(2) RL-based enhancement}, an emerging method for accurate evaluation~\citep{Judge-RL:1,Reas-GRM:2}. Another approach uses multi-agent interaction~\citep{liu2025syntheticsocraticdebatesexamining,ran2025bookworldnovelsinteractiveagent} or dialogue simulation~\citep{xu2024mindechoroleplayinglanguageagents,xiao2025humanizingmachinesrethinkingllm} to enhance judge capabilities through collaboration.

RLVR's success~\citep{deepseek-r1} highlights the importance of verifiable reward signals in complex assessment scenarios. Recent GenRM studies show convergent approaches: enhancing reward verifiability through auxiliary information~\citep{GenRM-Ref:1,GenRM-Ref:2} and improving reasoning for better reward signals~\citep{Reas-GRM:1,Reas-GRM:3,Reas-GRM:4,GenPRM}. However, our research reveals underlying challenges in this evolving paradigm.

\subsection{Evaluation Bias of LLM judges}

While LLM-as-a-judge shows promise, prior work confirms it suffers from significant biases~\citep{Bias-All:1,Bias-All:3,bias-self-pre} including model-inherent (position~\citep{Bias-Position:1}, length~\citep{Bias-Length:1}, self-preference~\citep{Bias-Selfpreference:1}), social (gender/racial~\citep{Bias-Social:1}), and cognitive biases (bandwagon/beauty~\citep{Bias-Con:1,Bias-others:1}). 

However, these studies focus on simple scenarios, neglecting biases induced by complex auxiliary information—a gap we address. Recent work also finds reasoning models, despite superior accuracy, exhibit unexpected vulnerabilities~\citep{RLM-judge-bias:1,RLM-judge-bias:2}; we systematically investigate this paradox in complex evaluation.
\section{Task Definition}

\subsection{Complex Evaluation}

With the rapid development of LLMs, evaluation tasks have evolved from simple scenarios (e.g., summarization) to increasingly complex ones (e.g., multi-step reasoning), which fundamentally changed traditional evaluation paradigms. 

Through extensive observation of LLM judges in these complex assessment scenarios~\citep{verifier-math,heimdall,xverify}, we identify that prevailing approaches typically incorporate the following types of evaluation context:

\begin{itemize}[leftmargin=*]
    \item \textbf{Reference:} Correct answers or high-quality examples as a reference for assessments~\citep{Reference:2,GenRM-Ref:1,xu2025guessithinkingbenchmark}.
    \item \textbf{Rubrics:} Structured scoring criterias or step-by-step evaluation protocols~\citep{rubrics,rubrics-2,Judge-RL:1,huang-etal-2024-reliability}.
    \item \textbf{Background Knowledge:} Essential contextual information for grounded assessment~\citep{rubrics-writing,DBLP:journals/corr/abs-2404-12138,rubrics-rp}.
\end{itemize}

A typical evaluation paradigm may involve: (1) assimilating contextual knowledge, (2) executing rubric-guided analysis, and (3) conducting reference-based comparison. We define this evaluation paradigm as \textbf{complex evaluation}, involving both challenging scenarios and multifaceted auxiliary information utilization. 

Existing research has empirically demonstrated that this auxiliary-enhanced evaluation paradigm significantly improve evaluation accuracy~\citep{ref-and-rubrics}. However, the utilization of auxiliary information also introduces new challenges — while the overall accuracy of model assessment improves, the models become more prone to \textbf{exhibit new evaluation biases}, leading to performance degradation in some cases.

\subsection{Bias Source}

\begin{figure*}[t]
  \includegraphics[width=\linewidth]{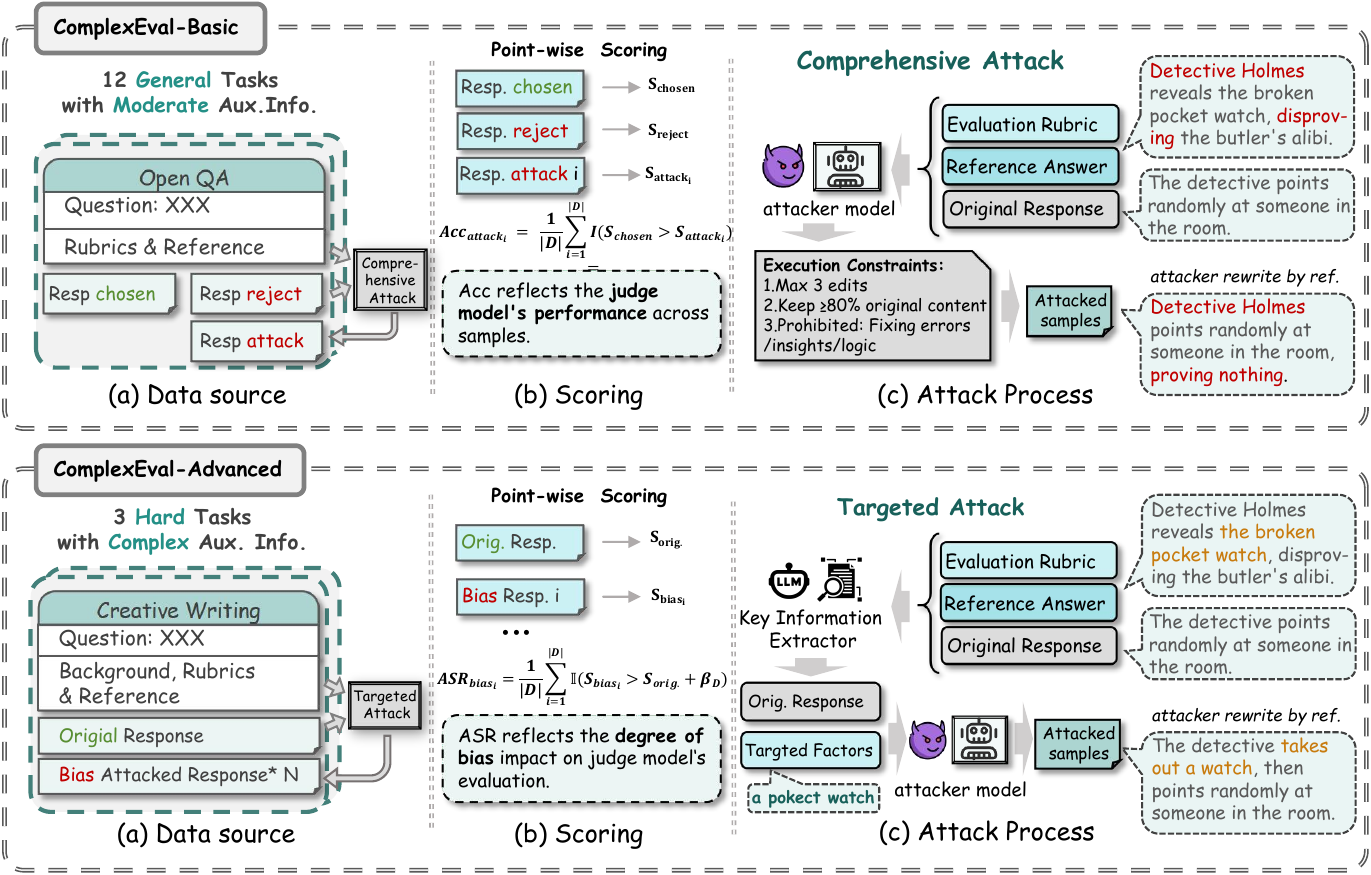} 
  \caption {The diagram illustrates the two core components of our dataset: 1) ComplexEval-Basic - Focused on 12 generic scenarios, employing Comprehensive Attack for macro-level bias exploration.
  2) ComplexEval-Advanced - Centered on 3 more complex scenarios, utilizing Targeted Attack for granular principle analysis}
  \label{pic.2}
\end{figure*}

\paragraph{Reference Bias}

While reference can help LLM judges conduct more consistent  assessment, it simultaneously introduces new evaluation biases, including but not limited to:

\begin{enumerate}[leftmargin=*]
    \item \textbf{\textit{Format Bias}:} Judges favor responses resembling reference in superficial features (e.g., structure, key words) over substantial quality.
    \item \textbf{\textit{Solution Fixation Bias}:}Judges often fixates on references, penalizing valid solutions that vary from them.
    \item \textbf{\textit{Stereotype Amplification Bias}:}The reference  may contain certain information (e.g., cultural characteristics) that could activate judges' inherent stereotypes, leading to evaluation biases. 
\end{enumerate}

\paragraph{Rubric Bias}

Using rubrics with various criteria can effectively enhance the judge's performance in unfamiliar/complex scenarios. However, such rubrics may also introduce cognitive biases,  including but not limited to:

\begin{enumerate}[leftmargin=*]
    \item \textbf{\textit{Criteria Loophole Bias}:}  For criteria not explicitly specified in the rubrics, judges relying on such rubrics will be more prone to overlook these features in scoring, thereby limiting their original evaluation capabilities.
    \item \textbf{\textit{Criteria Entanglement Bias}:} When presented with multiple criteria, judges tend to assign similar scores across dimensions (a "halo effect"), thereby undermining the rubrics'  advantage of enabling granular, differentiated evaluations. 
\end{enumerate}

Rubrics decompose evaluation into multiple dimensions, each targeting specific features. However, these  dimensions often overlap while still leaving gaps--directly corresponding to the two mentioned biases. Thus, while rubrics enable finer-grained assessment, their multi-dimensional complexity introduces new challenges.

\paragraph{Attention Limit Phenomenon} Complex evaluation may also introduce attention limit phenomenon—where judges exhibit performance bottlenecks when processing complex  auxiliary information. We will further investigate this phenomenon in subsequent experiments~\ref{5.3}

\section{\method}
\label{4}

\subsection{Adversarial Sample Generation}
\label{3.3}

To effectively quantitize these auxiliary information induced biases , we propose to construct a {dataset} via an adversarial {attack framework}, and design corresponding {metrics}. 

This framework synthesizes new samples by amplifying the bias-related features observed in the original dataset, following prior bias research~\citep{Bias-All:1}—by deploying an attacker model\footnote{GPT-4o-mini for comprehensive attack and DeepSeek-R1~\cite{deepseek-r1} for target attack.} to execute coordinated bias amplification.

\textbf{Comprehensive Attack (Instance-level):} The attacker processes each sample's  auxiliary information $A_i$ independently, generating unique adversarial examples through specific strategy:
   $  x'_i = \mathcal{A}_{CA}(x_i, A_i) $,
    where $\mathcal{A}_{CA}$ varies per instance.

\textbf{Targeted Attack (Type-level):} For bias category $k$, the attacker applies uniform extraction rules $R_k$ to all samples $\{x_j\}$, deriving type-homogeneous features $\phi_k^{(j)} = R_k(A_j)$. These features then generate standardized perturbations:
$ x'_j = \mathcal{A}_{TA}(x_j, \phi_k^{(j)}) $,
where $\phi_k^{(j)}$ shares identical structure across instances.

Attacks are strictly constrained in their modification extent to \textbf{avoid disrupting the original semantics.} Prompts for attack are shown in Appendix \ref{A.3}.


\subsection{Dataset Architecture}
\label{4.1}

To achieve both breadth and depth in bias analysis, we carefully design a two-tiered ComplexEval bench with distinct architectures. Table~\ref{1}. summarizes their key differences.

\begin{table}
\centering
\begin{tabular}{p{2.3cm}p{2.3cm}p{1.8cm}}
\hline
\textbf{Attribute} & \textbf{Basic} & \textbf{Advanced} \\
\hline
Data Format & $(Q, R_{c}, R_{r})$ & $(Q, R)$ \\
Info. Source & LLM-Generated & Human-Written \\
Complexity & Medium & High \\
\#Tasks & 12 domains & 3 domains \\
\#Samples & 1,056 & 400 \\
\hline
\end{tabular}
\caption{Key distinctions between ComplexEval tiers}
\label{1}
\end{table}

\subsubsection{\method-Basic}

\paragraph{Task \& Data Source} 

ComplexEval-Basic builds upon RMB-bench~\cite{RMB-bench}, a general benchmark for preference pairs used in reward modeling. Through domain-specific sampling (Appendix~\ref{A.1}), we construct 1,056 samples $(Q, R_{c}, R_{r})$ across 12 diverse scenarios (code, open-QA, etc.).

\paragraph{Auxiliary Information Synthesis}

Since original samples lack required auxiliary information, we adopt two augmentation steps to generate reference answers and rubrics (Appendix~\ref{A.2}). 

\paragraph{Attack Process}

Employing the Comprehensive Attack strategy (Section~\ref{3.3}), we systematically perturb the weaker responses $R_{r}$ in all preference pairs through both $A_{ref}$ and $A_{rubric}$ modifications:
\[
R'_{r} = \mathcal{A}_{GA}(R_{r}, A), \quad A \in \{A_{ref}, A_{rubric}\}
\]


\subsubsection{ComplexEval-Advanced}

\paragraph{Tasks \& Data Source} 

ComplexEval-Advanced focuses on three challenging domains requiring sophisticated evaluation. Through Random sampling (Appendix~\ref{A.1}), we curate 400 high-complexity single-response samples $(Q, R)$:

\begin{itemize}[leftmargin=*]
    \item \textbf{\textit{Mathematical Reasoning}}: mathematical reasoning from ProcessBench's Olympiad competition data\citep{ProcessBench}, with Official Olympiad solutions as $A_{ref}$.
    \item \textbf{\textit{Creative Writing}}: creative writing from WritingBench\citep{WritingBench} with multi-dimensional scoring systems as  $A_{rubric}$ and gold stories as $A_{ref}$.
    \item \textbf{\textit{Role-Playing}}: The most challenging scenario from the \textsc{CoSER} dataset\citep{CoSER}, featuring canonical dialogues $A_{ref}$ augmented with comprehensive character backgrounds and otherauxiliary information, evaluated via professional dramatic consistency rubrics $A_{rubric}$.
\end{itemize}

\paragraph{Attack Process}

 We apply Target Attack (Section~\ref{3.3}) to systematically probe five bias mechanisms. For each sample $(Q_i, R_i)$, perturbations are generated through type-consistent transformations:
\[R'_i = \mathcal{A}_{TA}(R_i, \phi_k), \quad \phi_k = R_k(A_i)\]


\subsection{Metric}
\label{4.2}

To ensure consistency across ComplexEval's tiered structure (Basic: preference pairs; Advanced: single responses), we implement a unified point-wise scoring approach based on recent GRM methods~\citep{Reas-GRM:1}.

For any query $Q$ with auxiliary information $A$, the evaluation process computes:$S = M(Q, R, A)$, where $M$ denotes the judge model's scoring function. 


\paragraph{For ComplexEval-Basic} We quantify judgment correctness using pairwise comparisons. While point-wise evaluation is operationally convenient, it requires special handling of score ties—overly strict equality criteria would inflate error rates. We therefore introduce a tolerance threshold $\theta=0.5$ to define ambiguous judgments when scores are functionally equivalent(Appendix~\ref{D}):
\[
J = 
\begin{cases} 
1, & S_c^{ref} > S_r^{ref} + \theta \\
0.5, & |S_c^{ref} - S_r^{ref}| \leq \theta \quad \\
0, & S_c^{ref} < S_r^{ref} - \theta
\end{cases}
\]
This $\theta$ value was empirically determined to balance sensitivity and specificity across our 12 scenarios. We then define two complementary metrics:
\begin{align*}
\text{ACC} &= \frac{1}{|D|}\sum_{i=1}^{|D|} \mathbb{I}(\hat{J}_i = J_i^*)  \\
\text{RR} &= \frac{1}{|D|}\sum_{i=1}^{|D|} \mathbb{I}(\hat{J}_i = J_i) \quad 
\end{align*}
where $J^*$ denotes ground-truth preferences and $\hat{J}$ represents judgments under bias-inducing conditions, $|D|$ denotes the dataset size.

\paragraph{For ComplexEval-Advanced}, the absence of preference labels prevents direct accuracy measurement. Instead, we employ attack success rate (ASR) as our primary bias metric, quantifying how frequently adversarial perturbations ($R \rightarrow R'$) induce significant score changes. This approach aligns with security evaluation paradigms where:

\begin{itemize}[leftmargin=*]
    \item \textit{Mathematical Reasoning}: An attack succeeds when the model correctly locates errors originally ($\text{Loc}_{\text{orig}} = \text{Loc}_{\text{gt}}$) but fails after perturbation ($\text{Loc}_{\text{adv}} = \neq  \text{Loc}_{\text{gt}}$):
    \[
    \text{Win} = \mathbb{I}(\text{Loc}_{\text{orig}} = \text{Loc}_{\text{gt}} \land \text{Loc}_{\text{adv}} \neq  \text{Loc}_{\text{gt}})
    \]
    
    \item \textit{Creative Writing/Role-Playing}: Success requires exceeding domain-specific improvement thresholds ($\beta_{\text{role}}=10$, $\beta_{\text{writing}}=0.5$), calibrated to each rubric's score distribution:
    \[
    \text{Win} = \mathbb{I}(S_{\text{adv}} - S_{\text{orig}} \geq \beta)
    \]
\end{itemize}

The overall ASR then provides a standardized bias measure:
\[
\text{ASR} = \frac{1}{N}\sum_{i=1}^N \text{Win}_i \quad 
\]
where $N$ represents attackable samples number.

\definecolor{darkgreen}{rgb}{0.13, 0.55, 0.13}

\begin{table*}[htbp]
\centering
\small
\begin{minipage}[t]{0.48\textwidth}
\centering
\setlength{\tabcolsep}{7pt}
\begin{tabular}{l|ccc}
\toprule
\multirow{2}{*}{Model} & \multicolumn{3}{c}{\textit{original samples}} \\
\cmidrule(lr){2-4} 
& $\text{Acc}_\text{orig.}$ & $\text{Acc}_\text{ref}$ & $\text{Acc}_\text{rubrics}$ \\
\midrule
\textit{General model} & & & \\
qwen2.5-32b & 
67.14 & 
69.81\textsubscript{\textcolor{darkgreen}{+2.67}} & 
66.89\textsubscript{\textcolor{red}{-0.25}}  \\
deepseek-v3 & 
69.24 & 
73.98\textsubscript{\textcolor{darkgreen}{+4.74}} & 
69.78\textsubscript{\textcolor{darkgreen}{+0.54}} \\
gpt-4o-mini & 
67.52 & 
70.64\textsubscript{\textcolor{darkgreen}{+3.12}} & 
70.80\textsubscript{\textcolor{darkgreen}{+3.28}}  \\

\midrule
\textit{Reasoning Model} & & & \\
qwq-32b & 
68.62 & 
72.62\textsubscript{\textcolor{darkgreen}{+4.00}} & 
70.30\textsubscript{\textcolor{darkgreen}{+1.68}}  \\
deepseek-r1 & 
72.32 & 
72.89\textsubscript{\textcolor{darkgreen}{+0.57}} & 
72.72\textsubscript{\textcolor{darkgreen}{+0.40}} \\
o4-mini & 
73.58 & 
76.33\textsubscript{\textcolor{darkgreen}{+2.75}} & 
74.15\textsubscript{\textcolor{darkgreen}{+0.57}} \\
\bottomrule
\end{tabular}
\caption{\textbf{ComplexEval-Basic:}Accuracy (\%) on original samples with auxiliary information (ref/rubrics) shows \textcolor{darkgreen}{improved} LLM judges' performance}
\label{tab:1}
\end{minipage}
\hfill
\begin{minipage}[t]{0.5\textwidth}
\centering
\setlength{\tabcolsep}{5pt}
\begin{tabular}{l|cc|cc}
\toprule
\multirow{2}{*}{Model} & \multicolumn{2}{c}{\textit{ref attack}} & \multicolumn{2}{c}{\textit{rubrics attack}} \\
\cmidrule(lr){2-3} \cmidrule(lr){4-5} 
 & $\text{Acc}_\text{orig.}$ & $\text{Acc}_\text{ref}$ & $\text{Acc}_\text{orig.}$ & $\text{Acc}_\text{rubrics}$ \\
\midrule
\textit{General model} & & & & \\
qwen2.5-32b & 
60.94 & 
60.47\textsubscript{\textcolor{red}{-0.47}} & 
65.01 & 
64.42\textsubscript{\textcolor{red}{-0.59}}\\
deepseek-v3 & 
59.72 & 
58.06\textsubscript{\textcolor{red}{-1.66}} & 
67.20 & 
66.32\textsubscript{\textcolor{red}{-0.88}}  \\
gpt-4o-mini & 
56.72 & 
56.53\textsubscript{\textcolor{red}{-0.19}} & 
63.64 & 
63.53\textsubscript{\textcolor{red}{-0.11}} \\

\midrule
\textit{Reasoning Model} & & & & \\
qwq-32b & 
62.10 & 
60.95\textsubscript{\textcolor{red}{-1.15}} & 
68.67 & 
66.51\textsubscript{\textcolor{red}{\textbf{-2.16}}} \\
deepseek-r1 & 
64.31 & 
59.81\textsubscript{\textcolor{red}{\textbf{-4.50}}} & 
71.47 & 
69.26\textsubscript{\textcolor{red}{\textbf{-2.21}}}  \\
o4-mini & 
64.90 & 
63.90\textsubscript{\textcolor{red}{\textbf{-1.00}}} & 
71.25 & 
\textbf{71.11}\textsubscript{\textcolor{red}{-0.14}} \\
\bottomrule
\end{tabular}
\caption{\textbf{ComplexEval-Basic:}Accuracy (\%) on attacked samples reveals auxiliary information instead \textcolor{red}{harms} LLM judges' performance}
\label{tab:3}
\end{minipage}
\end{table*}

\begin{table*}[htbp]
\centering
\small\setlength{\tabcolsep}{2.4pt}
\begin{tabular}{l|ccc|cccccc|ccccc}
\toprule
\multirow{2}{*}{Model} & \multicolumn{3}{c|}{Process Bench} & \multicolumn{6}{c|}{CoSER} & \multicolumn{5}{c}{Writing Bench} \\
\cmidrule(lr){2-4} \cmidrule(lr){5-10} \cmidrule(lr){11-15}
& Original & F.B. & S.F.B. & Original & F.B. & S.F.B. & S.A.B. & C.L.B. & C.E.B. & Original & F.B. & S.F.B. & C.L.B. & C.E.B. \\
\midrule
\textit{General model} & & & & & & & & & & & & & &  \\

gpt-4o-mini & 0.08 & \textbf{0.55} & 0.48 & 0.15  & 0.12 & 0.10 & 0.17 & \textbf{0.16} & 0.09 & 0.04 & 0.12& \textbf{0.19} & 0.15 & 0.19 \\
qwen2.5-32b & 0.27 & 0.40 & \textbf{0.44} & 0.15 & 0.14 & \textbf{0.22} & 0.18 & 0.14 & 0.17 & 0.15 & 0.19 & \textbf{0.45} & 0.27 & 0.29  \\
deepseek-v3 & 0.15 & 0.31 & \textbf{0.41} & 0.04 & 0.12 & \textbf{0.17} & 0.07 & 0.02 & 0.15 & 0.07 & 0.20 & \textbf{0.44} & 0.15 & 0.22 \\
\midrule
\textit{Reasoning Model} & & & & & & & & & & & & &  \\
qwq-32b & 0.11 & 0.27 & \textbf{0.47} & 0.15 & 0.33  & \textbf{0.45} & 0.18 & 0.30 & 0.28 & 0.19 & 0.28 & \textbf{0.52} & 0.29 & 0.38 \\
deepseek-r1 & 0.08 & 0.32 & \textbf{0.36} & 0.13 & 0.30  & \textbf{0.43} & 0.20 & 0.25 & 0.21 & 0.08 & 0.22 & \textbf{0.53} & 0.18 & 0.34 \\
\bottomrule
\end{tabular}

\caption{\textbf{ComplexEval-Advanced:} ASR comparison across task categories and bias types. Original represents non-attacked baseline samples. Abbreviations: F.B.=Format Bias, S.F.B.=Solution Fixation Bias, S.A.B.=Stereotype Amplification Bias, C.L.B.=Criteria Loophole Bias, C.E.B.=Criteria Entanglement Bias }
\label{tab:2}
\end{table*}

\section{Experiment}
\label{5}

\subsection{Setting}
To systematically compare reasoning-enhanced models against their general counterparts, we evaluate representative models following the paired-family methodology of~\citet{RLM-judge-bias:2}. Our selection includes: (1) Qwen2.5-32B~\citep{qwen2.5} vs QwQ-32B~\citep{qwq-32b}, (2) DeepSeek-V3~\citep{deepseek-v3} vs DeepSeek-R1~\citep{deepseek-r1}, (3) GPT-4o-mini~\citep{gpt-4} and (4) o4-mini~\citep{o4-mini}. This controlled comparison isolates the impact of reasoning capabilities on bias susceptibility while controlling for architectural similarities. All models are evaluated under identical prompting strategies (Appendix~\ref{A.3}).

\subsection{Results on \method-Basic} 

Using ComplexEval-Basic, we conduct a broad validation and phenomenological analysis of auxiliary information induced biases through comprehensive attack. Key findings from Table~2 reveal three fundamental insights:

\paragraph{Auxiliary information induced biases universally affect all scenarios and models.} As evidenced in Table~\ref{tab:1}\&~\ref{tab:3}, models demonstrate significantly degraded accuracy when processing attacked samples. More critically, while reference answers and rubrics improve evaluation accuracy on original samples (e.g., +4.74\% for DeepSeek-v3 with references), they paradoxically reduce performance on adversarial samples (-1.66\% for the same configuration). This counterintuitive inversion effect reveals the inherent dangers of auxiliary information induced biases.

\subsection{Results on \method-Advanced} 
\label{5.2}

Using ComplexEval-Advanced, we employ Targeted Attacks to isolate five hypothesized bias types through controlled perturbations (see Appendix~\ref{B.2} for attack templates). 

For our proposed bias typology, Table~\ref{tab:2} presents detailed attack success rates across evaluation scenarios. Notably, in certain scenarios (e.g., mathematical reasoning without evaluation rubrics), some types of auxiliary information are inherently unavailable, thus limiting the bias experiments across datasets.

\subsubsection{Reference Biases} 

\textbf{Format Bias shows models' overreliance on reference answer structure}, where similar responses get inflated scores regardless of quality. We confirm this by transplanting reference styles while keeping content quality constant. As Table~\ref{tab:2} (FB) shows, this bias persists across domains, especially in mathematical reasoning, where it achieves the highest attack success rate. We link this to models overvaluing formulaic expressions (e.g., "therefore", "let x be") as quality indicators.

\textbf{Solution Fixation Bias reflects models' cognitive rigidity in utilizing reference answer (Table \ref{tab:2})}. This bias is most severe when models use auxiliary cues, with QwQ-32b showing >45\% ASR. Experiments reveal models disproportionately favor reference-aligned responses, even when semantically or methodologically inappropriate.

\textbf{Stereotype Amplification Bias occurs when auxiliary information exacerbates a model's inherent cognitive biases}, going beyond reference answers and evaluation criteria. Our approach extracts stereotypical traits from profiles and introduces context-contradictory but stereotype-aligned behaviors. As Table \ref{tab:2} shows, this bias strongly affects both reasoning and general models. While stereotypes reflect implicit biases, our results reveal their significant amplification when triggered by auxiliary information.

These effects manifest not only in reliance on incomplete information but persists even with perfectly correct references (e.g., valid reasoning chains in mathematical scenarios or canonical dialogues in role-playing). Current context-dependent utilization mechanisms distort models' implicit evaluation standards for both form and content, with reasoning-enhanced models showing greater susceptibility due to their extended reasoning chains' amplified dependence on auxiliary information (Appendix \ref{C}).

\subsubsection{Rubric Induced Biases}

\textbf{Criteria Loopholes Bias reveals models' limitations in exploiting incomplete evaluation frameworks, causing systematic assessment distortions.} This bias shows high prevalence in creative writing (e.g., Qwen2-32B achieves 27\% ASR) but remains negligible in role-playing scenarios due to methodological differences: role-playing uses operationally precise point-deduction scoring, while creative writing relies on holistic rubrics with systematically exploitable judgments (Appendix~\ref{A.3.1}).

\textbf{Criteria Entanglement Bias highlights models' tendency to assign similar scores across dimensions even when some are objectively weaker. }As shown in Table~\ref{tab:2}, this bias is strong in creative writing but reduced—though still present—in role-playing tasks. We attribute this mitigation to role-playing’s independent dimensional evaluation design, improving rule utilization. However, inherent dimension overlaps still cause notable coupling effects.

These effects are most evident in scoring-based multi-dimensional rubrics (e.g., creative writing). Dimension-independent evaluation can ease such biases but adds extra assessment overhead.

\subsection{In-Depth Analysis}

\begin{figure*}[t]
  \centering
  \begin{minipage}{0.48\textwidth}
    \centering
    \includegraphics[width=\linewidth]{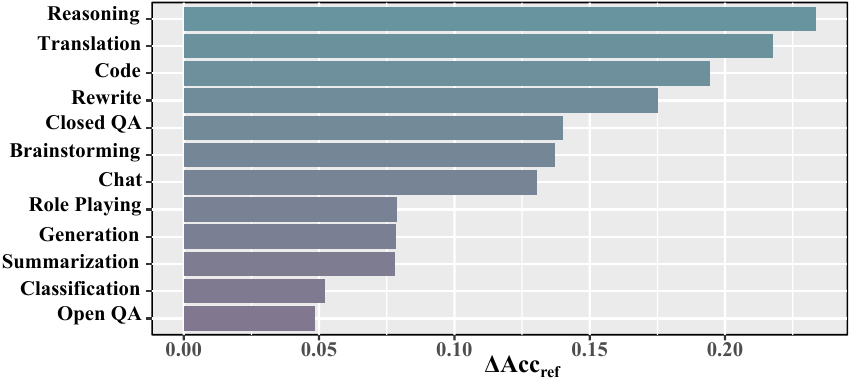}
    \caption{Cross-domain comparison of auxiliary information induced bias effects}
    \label{fig:4}
  \end{minipage}
  \hfill
  \begin{minipage}{0.48\textwidth}
    \centering
    \includegraphics[width=\linewidth]{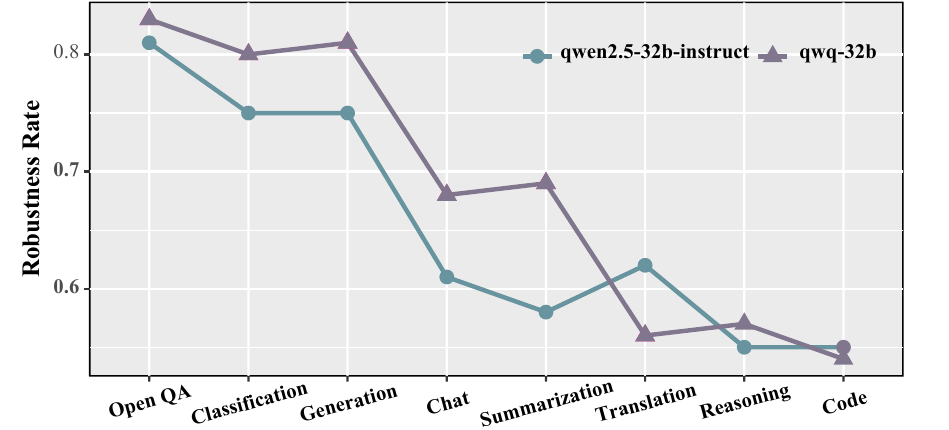}
    \caption{Robustness variation between reasoning and general models with increasing domain complexity}
    \label{fig:5}
  \end{minipage}
\end{figure*}

In systematic analysis based on the ComplexEval framework, we have also uncovered the following auxiliary information-related characteristics:


\textbf{1) Bias severity scales with evaluation scenario complexity.} As shown in Figure~\ref{fig:4}, more challenging scenarios—particularly those involving reasoning or open-ended humanities tasks like Reasoning, Code, Translation, and Rewrite—exhibit greater accuracy fluctuations than simpler tasks. Conversely, straightforward instruction-following tasks (e.g., Classification, Summarization) demonstrate higher robustness against such biases.

\textbf{2) Reasoning models show greater sensitivity to auxiliary information biases} As shown in Table~\ref{tab:1}\&~\ref{tab:3}, reasoning models suffer more severe degradation when evaluating biased samples with auxiliary information, with declines over twice those of general models (e.g., 3.35\% vs 1.27\% between deepseek-r1 and deepseek-v3). Our robustness analysis in Figure\ref{fig:5} highlights a key trend: while reasoning models remain more robust in simple scenarios, their advantage reverses in complex settings, often dropping to parity or below general models. This effect is even clearer in the fine-grained analysis ~\ref{tab:2}.

\begin{figure*}[t]
  \includegraphics[width=0.48\linewidth]{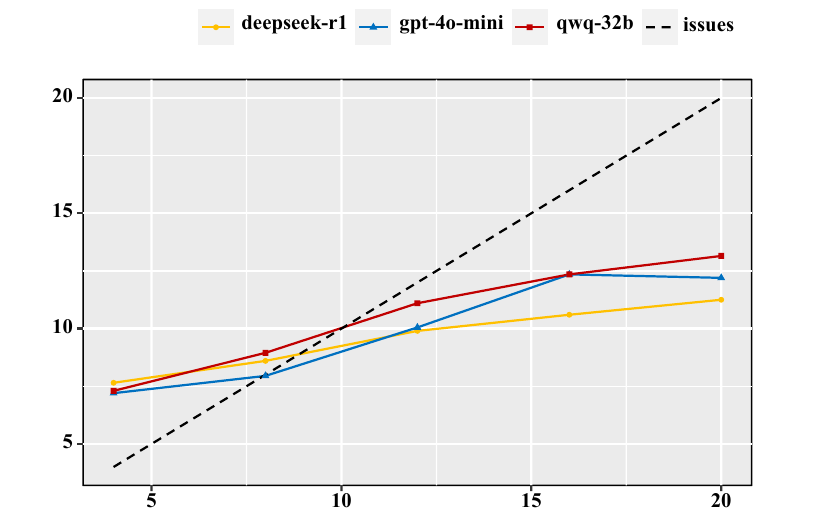} 
  \hfill
  \includegraphics[width=0.48\linewidth]{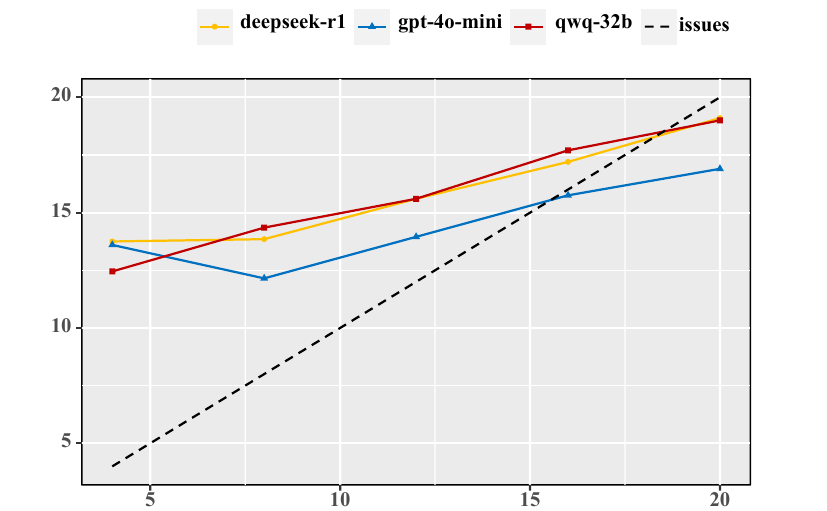}
  \caption {Evaluation of issue detection accuracy against ground truth (y=x).
Left: Multi-dimensional evaluation shows a hard ceiling (~15 issues) despite increasing actual issues; Right: Single-dimension evaluation removes the ceiling but introduces a floor (1-2 false detections).}
  \label{fig:6}
\end{figure*}

\label{5.3}
\textbf{3) Attention limit phenomenon reveals a critical flaw in multi-dimensional evaluation}: When judging others, models show predictable error patterns: their critiques remain within a fixed range. Using deduction-based scoring from the CoSER dataset, we quantify this effect on role-play dialogue samples containing 20 distinct issues (Figure~\ref{fig:6}), each linked to a fine-grained criterion. Results reveal an upper limit on the number of issues identifiable as dimensionality increases. As shown in Figure~\ref{fig:6} (right), dimension-wise evaluation mitigates this limit but also exposes a lower bound: models still flag 1–2 issues even when none exist.

\textbf{4) Evaluation models develop biases primarily due to three interconnected causes.} Our case studies reveal three main sources: (a) over-reliance on reference answers, (b) underutilization of structured rubrics, and (c) amplification of latent flaws in auxiliary information. Models often depend too heavily on references—even when those references are correct—leading to distorted implicit standards, especially the reasoning models(Appendix \ref{C}). At the same time, they make limited use of rubrics, which causes attention overload and blurred evaluation criteria. These issues reinforce one another, as hidden flaws in references or guidelines become amplified during assessment.

Our findings highlight the need to improve models' information utilization in future evaluation system design. Recent work has explored directions such as selectively using verifiable reward signals to enhance reliability~\citep{agentRM} or generating instance-level evaluation rules~\citep{Reas-GRM:1,Judge-RL:1}. Our study further shows that in complex scenarios, current approaches relying on auxiliary information introduce significant biases due to models' limited capacity for processing it.


\section{Conclusion}

In summary, this study formally identifies and validates a category of auxiliary information induced biases previously unexplored in prior research, demonstrating how these biases undermine the reliability of large language model evaluations in complex scenarios. We further conduct fine-grained analysis of five specific bias types, revealing their essential nature stems from models' limited capacity to effectively utilize auxiliary information. Through this work, we alert the research community to the critical severity of evaluation biases in complex scenarios, and aim to provide foundational insights for advancing LLM-as-a-judge capabilities toward more sophisticated and general-purpose evaluation settings.

\section*{Limitations}

While our work systematically characterizes auxiliary information induced biases, two important limitations warrant discussion:

\textbf{Scope of Bias Coverage: }Our study primarily focuses on biases induced by two mainstream forms of auxiliary information—reference answers and scoring rubrics. While we also explore the influence of other auxiliary signals (e.g., Stereo Amplification Bias), our coverage remains limited. Additional types of information may introduce novel bias patterns not yet captured in ComplexEval. For instance, multi-modal reference materials or interactive tool-use scenarios could yield distinct mechanisms of bias that warrant further investigation.

\textbf{Mitigation Strategies:} Although we identify the dual origin of these biases (model limitations and information quality) and propose initial mitigation directions (Section 6), developing comprehensive solutions remains an open challenge. Interventions like controlled information integration require rigorous validation across diverse evaluation frameworks. Moreover, there may be fundamental trade-offs between bias reduction and maintaining evaluation accuracy—a critical area for future theoretical and empirical work.

These limitations, however, reflect intentional scope boundaries rather than oversight. Our focused approach enables deeper analysis of high-impact bias categories while providing a methodological foundation (ComplexEval) for extending this research to additional types of auxiliary information and mitigation strategies.

\section*{Ethics Statements}

In this work, our bias attacks focus solely on fair evaluation. The modifications and enhancements applied to samples in the constructed dataset do not involve any ethical concerns. Furthermore, this paper proposes a method for constructing auxiliary information-based adversarial samples to influence the fairness evaluation of large models. It is important to emphasize that the purpose of this method is to alert the community to the significance of such biases and to explore potential improvements for LLM judges—not to encourage the use of this approach for conducting bias attacks.

\section*{Acknowledgments}

This work is supported by the Chinese NSF Major
Research Plan (No.92270121), General Program (No.62572129).

\bibliography{reference}

\begin{thebibliography}{74}
\providecommand{\natexlab}[1]{#1}

\bibitem[{Bai et~al.(2022)Bai, Kadavath, Kundu, Askell, Kernion, Jones, Chen, Goldie, Mirhoseini, McKinnon, Chen, Olsson, Olah, Hernandez, Drain, Ganguli, Li, Tran-Johnson, Perez, Kerr, Mueller, Ladish, Landau, Ndousse, Lukosuite, Lovitt, Sellitto, Elhage, Schiefer, Mercado, DasSarma, Lasenby, Larson, Ringer, Johnston, Kravec, Showk, Fort, Lanham, Telleen-Lawton, Conerly, Henighan, Hume, Bowman, Hatfield-Dodds, Mann, Amodei, Joseph, McCandlish, Brown, and Kaplan}]{Judge-Old-Work:1}
Yuntao Bai, Saurav Kadavath, Sandipan Kundu, Amanda Askell, Jackson Kernion, Andy Jones, Anna Chen, Anna Goldie, Azalia Mirhoseini, Cameron McKinnon, Carol Chen, Catherine Olsson, Christopher Olah, Danny Hernandez, Dawn Drain, Deep Ganguli, Dustin Li, Eli Tran-Johnson, Ethan Perez, and 32 others. 2022.
\newblock \href {https://arxiv.org/abs/2212.08073} {Constitutional ai: Harmlessness from ai feedback}.
\newblock \emph{Preprint}, arXiv:2212.08073.

\bibitem[{Brown et~al.(2020)Brown, Mann, Ryder, Subbiah, Kaplan, Dhariwal, Neelakantan, Shyam, Sastry, Askell, Agarwal, Herbert-Voss, Krueger, Henighan, Child, Ramesh, Ziegler, Wu, Winter, Hesse, Chen, Sigler, Litwin, Gray, Chess, Clark, Berner, McCandlish, Radford, Sutskever, and Amodei}]{LLM:1}
Tom Brown, Benjamin Mann, Nick Ryder, Melanie Subbiah, Jared~D Kaplan, Prafulla Dhariwal, Arvind Neelakantan, Pranav Shyam, Girish Sastry, Amanda Askell, Sandhini Agarwal, Ariel Herbert-Voss, Gretchen Krueger, Tom Henighan, Rewon Child, Aditya Ramesh, Daniel Ziegler, Jeffrey Wu, Clemens Winter, and 12 others. 2020.
\newblock \href {https://proceedings.neurips.cc/paper_files/paper/2020/file/1457c0d6bfcb4967418bfb8ac142f64a-Paper.pdf} {Language models are few-shot learners}.
\newblock In \emph{Advances in Neural Information Processing Systems}, volume~33, pages 1877--1901. Curran Associates, Inc.

\bibitem[{Chan et~al.(2024)Chan, Chen, Su, Yu, Xue, Zhang, Fu, and Liu}]{judge:5}
Chi{-}Min Chan, Weize Chen, Yusheng Su, Jianxuan Yu, Wei Xue, Shanghang Zhang, Jie Fu, and Zhiyuan Liu. 2024.
\newblock \href {https://openreview.net/forum?id=FQepisCUWu} {Chateval: Towards better llm-based evaluators through multi-agent debate}.
\newblock In \emph{The Twelfth International Conference on Learning Representations, {ICLR} 2024, Vienna, Austria, May 7-11, 2024}. OpenReview.net.

\bibitem[{Chen et~al.(2025{\natexlab{a}})Chen, Yu, Wang, Zhang, Tang, Xiong, Li, Yang, and Li}]{xverify}
Ding Chen, Qingchen Yu, Pengyuan Wang, Wentao Zhang, Bo~Tang, Feiyu Xiong, Xinchi Li, Minchuan Yang, and Zhiyu Li. 2025{\natexlab{a}}.
\newblock xverify: Efficient answer verifier for reasoning model evaluations.
\newblock \emph{arXiv preprint arXiv:2504.10481}.

\bibitem[{Chen et~al.(2024{\natexlab{a}})Chen, Chen, Liu, Jiang, and Wang}]{Bias-others:1}
Guiming Chen, Shunian Chen, Ziche Liu, Feng Jiang, and Benyou Wang. 2024{\natexlab{a}}.
\newblock \href {https://aclanthology.org/2024.emnlp-main.474} {Humans or llms as the judge? {A} study on judgement bias}.
\newblock In \emph{Proceedings of the 2024 Conference on Empirical Methods in Natural Language Processing, {EMNLP} 2024, Miami, FL, USA, November 12-16, 2024}, pages 8301--8327.

\bibitem[{Chen et~al.(2024{\natexlab{b}})Chen, Wang, Xu, Yuan, Zhang, Shi, Xie, Li, Yang, Zhu, Chen, Li, Chen, Hu, Wu, Ren, Fu, and Xiao}]{chen2024personapersonalizationsurveyroleplaying}
Jiangjie Chen, Xintao Wang, Rui Xu, Siyu Yuan, Yikai Zhang, Wei Shi, Jian Xie, Shuang Li, Ruihan Yang, Tinghui Zhu, Aili Chen, Nianqi Li, Lida Chen, Caiyu Hu, Siye Wu, Scott Ren, Ziquan Fu, and Yanghua Xiao. 2024{\natexlab{b}}.
\newblock \href {https://arxiv.org/abs/2404.18231} {From persona to personalization: A survey on role-playing language agents}.
\newblock \emph{Preprint}, arXiv:2404.18231.

\bibitem[{Chen et~al.(2025{\natexlab{b}})Chen, Hu, Zou, Wu, Wang, Hooi, and He}]{Reas-GRM:2}
Nuo Chen, Zhiyuan Hu, Qingyun Zou, Jiaying Wu, Qian Wang, Bryan Hooi, and Bingsheng He. 2025{\natexlab{b}}.
\newblock Judgelrm: Large reasoning models as a judge.
\newblock \emph{arXiv preprint arXiv:2504.00050}.

\bibitem[{Chen et~al.(2025{\natexlab{c}})Chen, Li, Wang, Jin, Qian, Wang, Wang, Zhang, Zhang, Zhang, Tong, and Ji}]{Reas-GRM:3}
Xiusi Chen, Gaotang Li, Ziqi Wang, Bowen Jin, Cheng Qian, Yu~Wang, Hongru Wang, Yu~Zhang, Denghui Zhang, Tong Zhang, Hanghang Tong, and Heng Ji. 2025{\natexlab{c}}.
\newblock \href {https://arxiv.org/abs/2505.02387} {Rm-r1: Reward modeling as reasoning}.
\newblock \emph{Preprint}, arXiv:2505.02387.

\bibitem[{Cobbe et~al.(2021)Cobbe, Kosaraju, Bavarian, Chen, Jun, Kaiser, Plappert, Tworek, Hilton, Nakano, Hesse, and Schulman}]{verifier-math}
Karl Cobbe, Vineet Kosaraju, Mohammad Bavarian, Mark Chen, Heewoo Jun, Lukasz Kaiser, Matthias Plappert, Jerry Tworek, Jacob Hilton, Reiichiro Nakano, Christopher Hesse, and John Schulman. 2021.
\newblock \href {https://arxiv.org/abs/2110.14168} {Training verifiers to solve math word problems}.
\newblock \emph{CoRR}, abs/2110.14168.

\bibitem[{DeepSeek-AI et~al.(2025{\natexlab{a}})DeepSeek-AI, Guo, Yang, Zhang, Song, Zhang, Xu, Zhu, Ma, Wang, Bi, Zhang, Yu, Wu, Wu, Gou, Shao, Li, Gao, Liu, Xue, Wang, Wu, Feng, Lu, Zhao, Deng, Zhang, Ruan, Dai, Chen, Ji, Li, Lin, Dai, Luo, Hao, Chen, Li, Zhang, Bao, Xu, Wang, Ding, Xin, Gao, Qu, Li, Guo, Li, Wang, Chen, Yuan, Qiu, Li, Cai, Ni, Liang, Chen, Dong, Hu, Gao, Guan, Huang, Yu, Wang, Zhang, Zhao, Wang, Zhang, Xu, Xia, Zhang, Zhang, Tang, Li, Wang, Li, Tian, Huang, Zhang, Wang, Chen, Du, Ge, Zhang, Pan, Wang, Chen, Jin, Chen, Lu, Zhou, Chen, Ye, Wang, Yu, Zhou, Pan, Li, Zhou, Wu, Ye, Yun, Pei, Sun, Wang, Zeng, Zhao, Liu, Liang, Gao, Yu, Zhang, Xiao, An, Liu, Wang, Chen, Nie, Cheng, Liu, Xie, Liu, Yang, Li, Su, Lin, Li, Jin, Shen, Chen, Sun, Wang, Song, Zhou, Wang, Shan, Li, Wang, Wei, Zhang, Xu, Li, Zhao, Sun, Wang, Yu, Zhang, Shi, Xiong, He, Piao, Wang, Tan, Ma, Liu, Guo, Ou, Wang, Gong, Zou, He, Xiong, Luo, You, Liu, Zhou, Zhu, Xu, Huang, Li, Zheng, Zhu, Ma, Tang, Zha, Yan, Ren, Ren, Sha, Fu,
  Xu, Xie, Zhang, Hao, Ma, Yan, Wu, Gu, Zhu, Liu, Li, Xie, Song, Pan, Huang, Xu, Zhang, and Zhang}]{deepseek-r1}
DeepSeek-AI, Daya Guo, Dejian Yang, Haowei Zhang, Junxiao Song, Ruoyu Zhang, Runxin Xu, Qihao Zhu, Shirong Ma, Peiyi Wang, Xiao Bi, Xiaokang Zhang, Xingkai Yu, Yu~Wu, Z.~F. Wu, Zhibin Gou, Zhihong Shao, Zhuoshu Li, Ziyi Gao, and 181 others. 2025{\natexlab{a}}.
\newblock \href {https://arxiv.org/abs/2501.12948} {Deepseek-r1: Incentivizing reasoning capability in llms via reinforcement learning}.
\newblock \emph{Preprint}, arXiv:2501.12948.

\bibitem[{DeepSeek-AI et~al.(2025{\natexlab{b}})DeepSeek-AI, Liu, Feng, Xue, Wang, Wu, Lu, Zhao, Deng, Zhang, Ruan, Dai, Guo, Yang, Chen, Ji, Li, Lin, Dai, Luo, Hao, Chen, Li, Zhang, Bao, Xu, Wang, Zhang, Ding, Xin, Gao, Li, Qu, Cai, Liang, Guo, Ni, Li, Wang, Chen, Chen, Yuan, Qiu, Li, Song, Dong, Hu, Gao, Guan, Huang, Yu, Wang, Zhang, Xu, Xia, Zhao, Wang, Zhang, Li, Wang, Zhang, Zhang, Tang, Li, Tian, Huang, Wang, Zhang, Wang, Zhu, Chen, Du, Chen, Jin, Ge, Zhang, Pan, Wang, Xu, Zhang, Chen, Li, Lu, Zhou, Chen, Wu, Ye, Ye, Ma, Wang, Zhou, Yu, Zhou, Pan, Wang, Yun, Pei, Sun, Xiao, Zeng, Zhao, An, Liu, Liang, Gao, Yu, Zhang, Li, Jin, Wang, Bi, Liu, Wang, Shen, Chen, Zhang, Chen, Nie, Sun, Wang, Cheng, Liu, Xie, Liu, Yu, Song, Shan, Zhou, Yang, Li, Su, Lin, Li, Wang, Wei, Zhu, Zhang, Xu, Xu, Huang, Li, Zhao, Sun, Li, Wang, Yu, Zheng, Zhang, Shi, Xiong, He, Tang, Piao, Wang, Tan, Ma, Liu, Guo, Wu, Ou, Zhu, Wang, Gong, Zou, He, Zha, Xiong, Ma, Yan, Luo, You, Liu, Zhou, Wu, Ren, Ren, Sha, Fu, Xu, Huang, Zhang, Xie,
  Zhang, Hao, Gou, Ma, Yan, Shao, Xu, Wu, Zhang, Li, Gu, Zhu, Liu, Li, Xie, Song, Gao, and Pan}]{deepseek-v3}
DeepSeek-AI, Aixin Liu, Bei Feng, Bing Xue, Bingxuan Wang, Bochao Wu, Chengda Lu, Chenggang Zhao, Chengqi Deng, Chenyu Zhang, Chong Ruan, Damai Dai, Daya Guo, Dejian Yang, Deli Chen, Dongjie Ji, Erhang Li, Fangyun Lin, Fucong Dai, and 181 others. 2025{\natexlab{b}}.
\newblock \href {https://arxiv.org/abs/2412.19437} {Deepseek-v3 technical report}.
\newblock \emph{Preprint}, arXiv:2412.19437.

\bibitem[{Gao et~al.(2023)Gao, Ruan, Sun, Yin, Yang, and Wan}]{Judge-Old-Work:2}
Mingqi Gao, Jie Ruan, Renliang Sun, Xunjian Yin, Shiping Yang, and Xiaojun Wan. 2023.
\newblock \href {https://arxiv.org/abs/2304.02554} {Human-like summarization evaluation with chatgpt}.
\newblock \emph{Preprint}, arXiv:2304.02554.

\bibitem[{Gilardi et~al.(2023)Gilardi, Alizadeh, and Kubli}]{judge:2}
Fabrizio Gilardi, Meysam Alizadeh, and Ma{\"e}l Kubli. 2023.
\newblock Chatgpt outperforms crowd workers for text-annotation tasks.
\newblock \emph{Proceedings of the National Academy of Sciences}, 120(30):e2305016120.

\bibitem[{Gu et~al.(2024)Gu, Jiang, Shi, Tan, Zhai, Xu, Li, Shen, Ma, Liu et~al.}]{judge-survey:2}
Jiawei Gu, Xuhui Jiang, Zhichao Shi, Hexiang Tan, Xuehao Zhai, Chengjin Xu, Wei Li, Yinghan Shen, Shengjie Ma, Honghao Liu, and 1 others. 2024.
\newblock A survey on llm-as-a-judge.
\newblock \emph{arXiv preprint arXiv:2411.15594}.

\bibitem[{Huang et~al.(2024)Huang, Jiao, Lam, Li, Wang, and Lyu}]{huang-etal-2024-reliability}
Jen-tse Huang, Wenxiang Jiao, Man~Ho Lam, Eric~John Li, Wenxuan Wang, and Michael Lyu. 2024.
\newblock \href {https://doi.org/10.18653/v1/2024.emnlp-main.354} {On the reliability of psychological scales on large language models}.
\newblock In \emph{Proceedings of the 2024 Conference on Empirical Methods in Natural Language Processing}, pages 6152--6173, Miami, Florida, USA. Association for Computational Linguistics.

\bibitem[{Huang et~al.(2025)Huang, Zhuang, Lu, Qin, Xu, Zhao, Peng, Hu, Shen, Hu, Gu, Tu, Liu, Chen, Fu, Fan, Gu, Wang, Yang, Li, and Zhao}]{Rubrics-RL-2}
Zenan Huang, Yihong Zhuang, Guoshan Lu, Zeyu Qin, Haokai Xu, Tianyu Zhao, Ru~Peng, Jiaqi Hu, Zhanming Shen, Xiaomeng Hu, Xijun Gu, Peiyi Tu, Jiaxin Liu, Wenyu Chen, Yuzhuo Fu, Zhiting Fan, Yanmei Gu, Yuanyuan Wang, Zhengkai Yang, and 2 others. 2025.
\newblock \href {https://arxiv.org/abs/2508.12790} {Reinforcement learning with rubric anchors}.
\newblock \emph{Preprint}, arXiv:2508.12790.

\bibitem[{Kim et~al.(2024)Kim, Shin, Choi, Jang, Longpre, Lee, Yun, Shin, Kim, Thorne, and Seo}]{ref-and-rubrics}
Seungone Kim, Jamin Shin, Yejin Choi, Joel Jang, Shayne Longpre, Hwaran Lee, Sangdoo Yun, Seongjin Shin, Sungdong Kim, James Thorne, and Minjoon Seo. 2024.
\newblock \href {https://openreview.net/forum?id=8euJaTveKw} {Prometheus: Inducing fine-grained evaluation capability in language models}.
\newblock In \emph{The Twelfth International Conference on Learning Representations, {ICLR} 2024, Vienna, Austria, May 7-11, 2024}. OpenReview.net.

\bibitem[{Kim et~al.(2025)Kim, Suk, Cho, Longpre, Kim, Yoon, Son, Cho, Shafayat, Baek, Park, Hwang, Jo, Cho, Shin, Lee, Oh, Lee, Ho, Joo, Ko, Lee, Chae, Shin, Jang, Ye, Lin, Welleck, Neubig, Lee, Lee, and Seo}]{judge-rules:3}
Seungone Kim, Juyoung Suk, Ji~Yong Cho, Shayne Longpre, Chaeeun Kim, Dongkeun Yoon, Guijin Son, Yejin Cho, Sheikh Shafayat, Jinheon Baek, Sue~Hyun Park, Hyeonbin Hwang, Jinkyung Jo, Hyowon Cho, Haebin Shin, Seongyun Lee, Hanseok Oh, Noah Lee, Namgyu Ho, and 13 others. 2025.
\newblock \href {https://aclanthology.org/2025.naacl-long.303/} {The {B}i{GG}en bench: A principled benchmark for fine-grained evaluation of language models with language models}.
\newblock In \emph{Proceedings of the 2025 Conference of the Nations of the Americas Chapter of the Association for Computational Linguistics: Human Language Technologies (Volume 1: Long Papers)}, pages 5877--5919, Albuquerque, New Mexico. Association for Computational Linguistics.

\bibitem[{Koo et~al.(2024)Koo, Lee, Raheja, Park, Kim, and Kang}]{Bias-Con:1}
Ryan Koo, Minhwa Lee, Vipul Raheja, Jong~Inn Park, Zae~Myung Kim, and Dongyeop Kang. 2024.
\newblock \href {https://doi.org/10.18653/V1/2024.FINDINGS-ACL.29} {Benchmarking cognitive biases in large language models as evaluators}.
\newblock In \emph{Findings of the Association for Computational Linguistics, {ACL} 2024, Bangkok, Thailand and virtual meeting, August 11-16, 2024}, pages 517--545. Association for Computational Linguistics.

\bibitem[{Krumdick et~al.(2025)Krumdick, Lovering, Reddy, Ebner, and Tanner}]{Reference:1}
Michael Krumdick, Charles Lovering, Varshini Reddy, Seth Ebner, and Chris Tanner. 2025.
\newblock No free labels: Limitations of llm-as-a-judge without human grounding.
\newblock \emph{arXiv preprint arXiv:2503.05061}.

\bibitem[{Lee et~al.(2024{\natexlab{a}})Lee, Phatale, Mansoor, Mesnard, Ferret, Lu, Bishop, Hall, Carbune, Rastogi, and Prakash}]{Judge-Old-Work:3}
Harrison Lee, Samrat Phatale, Hassan Mansoor, Thomas Mesnard, Johan Ferret, Kellie Lu, Colton Bishop, Ethan Hall, Victor Carbune, Abhinav Rastogi, and Sushant Prakash. 2024{\natexlab{a}}.
\newblock \href {https://openreview.net/forum?id=uydQ2W41KO} {{RLAIF} vs. {RLHF:} scaling reinforcement learning from human feedback with {AI} feedback}.
\newblock In \emph{Forty-first International Conference on Machine Learning, {ICML} 2024, Vienna, Austria, July 21-27, 2024}. OpenReview.net.

\bibitem[{Lee et~al.(2024{\natexlab{b}})Lee, Cai, Meng, Wang, and Wu}]{rubrics-writing}
Sanwoo Lee, Yida Cai, Desong Meng, Ziyang Wang, and Yunfang Wu. 2024{\natexlab{b}}.
\newblock \href {https://doi.org/10.18653/v1/2024.findings-emnlp.10} {Unleashing large language models' proficiency in zero-shot essay scoring}.
\newblock In \emph{Findings of the Association for Computational Linguistics: EMNLP 2024}, pages 181--198, Miami, Florida, USA. Association for Computational Linguistics.

\bibitem[{Li et~al.(2024)Li, Jiang, Huang, Beigi, Zhao, Tan, Bhattacharjee, Jiang, Chen, Wu et~al.}]{judge-survey:1}
Dawei Li, Bohan Jiang, Liangjie Huang, Alimohammad Beigi, Chengshuai Zhao, Zhen Tan, Amrita Bhattacharjee, Yuxuan Jiang, Canyu Chen, Tianhao Wu, and 1 others. 2024.
\newblock From generation to judgment: Opportunities and challenges of llm-as-a-judge.
\newblock \emph{arXiv preprint arXiv:2411.16594}.

\bibitem[{Li et~al.(2025{\natexlab{a}})Li, Sun, Huang, Zhong, Jiang, Han, Zhang, Wang, and Liu}]{bias-self-pre}
Dawei Li, Renliang Sun, Yue Huang, Ming Zhong, Bohan Jiang, Jiawei Han, Xiangliang Zhang, Wei Wang, and Huan Liu. 2025{\natexlab{a}}.
\newblock \href {https://doi.org/10.48550/ARXIV.2502.01534} {Preference leakage: {A} contamination problem in llm-as-a-judge}.
\newblock \emph{CoRR}, abs/2502.01534.

\bibitem[{Li et~al.(2025{\natexlab{b}})Li, Liu, Liu, Zhou, Diab, and Sap}]{li2025big5chatshapingllmpersonalities}
Wenkai Li, Jiarui Liu, Andy Liu, Xuhui Zhou, Mona Diab, and Maarten Sap. 2025{\natexlab{b}}.
\newblock \href {https://arxiv.org/abs/2410.16491} {Big5-chat: Shaping llm personalities through training on human-grounded data}.
\newblock \emph{Preprint}, arXiv:2410.16491.

\bibitem[{Liu et~al.(2025{\natexlab{a}})Liu, Song, Xiao, Zheng, Tjuatja, Borg, Diab, and Sap}]{liu2025syntheticsocraticdebatesexamining}
Jiarui Liu, Yueqi Song, Yunze Xiao, Mingqian Zheng, Lindia Tjuatja, Jana~Schaich Borg, Mona Diab, and Maarten Sap. 2025{\natexlab{a}}.
\newblock \href {https://arxiv.org/abs/2506.12657} {Synthetic socratic debates: Examining persona effects on moral decision and persuasion dynamics}.
\newblock \emph{Preprint}, arXiv:2506.12657.

\bibitem[{Liu et~al.(2024)Liu, Yang, Huang, Zhang, Huang, Wei, Deng, Sun, and Zhang}]{rubrics-2}
Yuxuan Liu, Tianchi Yang, Shaohan Huang, Zihan Zhang, Haizhen Huang, Furu Wei, Weiwei Deng, Feng Sun, and Qi~Zhang. 2024.
\newblock \href {https://aclanthology.org/2024.lrec-main.237} {Calibrating llm-based evaluator}.
\newblock In \emph{Proceedings of the 2024 Joint International Conference on Computational Linguistics, Language Resources and Evaluation, {LREC/COLING} 2024, 20-25 May, 2024, Torino, Italy}, pages 2638--2656. {ELRA} and {ICCL}.

\bibitem[{Liu et~al.(2025{\natexlab{b}})Liu, Wang, Xu, Ma, Ruan, Li, Liu, and Wu}]{Reas-GRM:1}
Zijun Liu, Peiyi Wang, Runxin Xu, Shirong Ma, Chong Ruan, Peng Li, Yang Liu, and Yu~Wu. 2025{\natexlab{b}}.
\newblock Inference-time scaling for generalist reward modeling.
\newblock \emph{arXiv preprint arXiv:2504.02495}.

\bibitem[{Mahan et~al.(2024)Mahan, Van~Phung, Rafailov, Blagden, Lile, Castricato, Fr{\"a}nken, Finn, and Albalak}]{GenRM:3}
Dakota Mahan, Duy Van~Phung, Rafael Rafailov, Chase Blagden, Nathan Lile, Louis Castricato, Jan-Philipp Fr{\"a}nken, Chelsea Finn, and Alon Albalak. 2024.
\newblock Generative reward models.
\newblock \emph{arXiv preprint arXiv:2410.12832}.

\bibitem[{{OpenAI}(2025)}]{o4-mini}
{OpenAI}. 2025.
\newblock \href {https://openai.com/index/o3-o4-mini-codex-system-card-addendum/} {Addendum to {O3} and {O4-mini} system card: {Codex}}.
\newblock Technical report.

\bibitem[{OpenAI et~al.(2024)OpenAI, Achiam, Adler, Agarwal, Ahmad, Akkaya, Aleman, Almeida, Altenschmidt, Altman, Anadkat, Avila, Babuschkin, Balaji, Balcom, Baltescu, Bao, Bavarian, Belgum, Bello, Berdine, Bernadett-Shapiro, Berner, Bogdonoff, Boiko, Boyd, Brakman, Brockman, Brooks, Brundage, Button, Cai, Campbell, Cann, Carey, Carlson, Carmichael, Chan, Chang, Chantzis, Chen, Chen, Chen, Chen, Chen, Chess, Cho, Chu, Chung, Cummings, Currier, Dai, Decareaux, Degry, Deutsch, Deville, Dhar, Dohan, Dowling, Dunning, Ecoffet, Eleti, Eloundou, Farhi, Fedus, Felix, Fishman, Forte, Fulford, Gao, Georges, Gibson, Goel, Gogineni, Goh, Gontijo-Lopes, Gordon, Grafstein, Gray, Greene, Gross, Gu, Guo, Hallacy, Han, Harris, He, Heaton, Heidecke, Hesse, Hickey, Hickey, Hoeschele, Houghton, Hsu, Hu, Hu, Huizinga, Jain, Jain, Jang, Jiang, Jiang, Jin, Jin, Jomoto, Jonn, Jun, Kaftan, Łukasz Kaiser, Kamali, Kanitscheider, Keskar, Khan, Kilpatrick, Kim, Kim, Kim, Kirchner, Kiros, Knight, Kokotajlo, Łukasz Kondraciuk,
  Kondrich, Konstantinidis, Kosic, Krueger, Kuo, Lampe, Lan, Lee, Leike, Leung, Levy, Li, Lim, Lin, Lin, Litwin, Lopez, Lowe, Lue, Makanju, Malfacini, Manning, Markov, Markovski, Martin, Mayer, Mayne, McGrew, McKinney, McLeavey, McMillan, McNeil, Medina, Mehta, Menick, Metz, Mishchenko, Mishkin, Monaco, Morikawa, Mossing, Mu, Murati, Murk, Mély, Nair, Nakano, Nayak, Neelakantan, Ngo, Noh, Ouyang, O'Keefe, Pachocki, Paino, Palermo, Pantuliano, Parascandolo, Parish, Parparita, Passos, Pavlov, Peng, Perelman, de~Avila Belbute~Peres, Petrov, de~Oliveira~Pinto, Michael, Pokorny, Pokrass, Pong, Powell, Power, Power, Proehl, Puri, Radford, Rae, Ramesh, Raymond, Real, Rimbach, Ross, Rotsted, Roussez, Ryder, Saltarelli, Sanders, Santurkar, Sastry, Schmidt, Schnurr, Schulman, Selsam, Sheppard, Sherbakov, Shieh, Shoker, Shyam, Sidor, Sigler, Simens, Sitkin, Slama, Sohl, Sokolowsky, Song, Staudacher, Such, Summers, Sutskever, Tang, Tezak, Thompson, Tillet, Tootoonchian, Tseng, Tuggle, Turley, Tworek, Uribe, Vallone,
  Vijayvergiya, Voss, Wainwright, Wang, Wang, Wang, Ward, Wei, Weinmann, Welihinda, Welinder, Weng, Weng, Wiethoff, Willner, Winter, Wolrich, Wong, Workman, Wu, Wu, Wu, Xiao, Xu, Yoo, Yu, Yuan, Zaremba, Zellers, Zhang, Zhang, Zhao, Zheng, Zhuang, Zhuk, and Zoph}]{gpt-4}
OpenAI, Josh Achiam, Steven Adler, Sandhini Agarwal, Lama Ahmad, Ilge Akkaya, Florencia~Leoni Aleman, Diogo Almeida, Janko Altenschmidt, Sam Altman, Shyamal Anadkat, Red Avila, Igor Babuschkin, Suchir Balaji, Valerie Balcom, Paul Baltescu, Haiming Bao, Mohammad Bavarian, Jeff Belgum, and 262 others. 2024.
\newblock \href {https://arxiv.org/abs/2303.08774} {Gpt-4 technical report}.
\newblock \emph{Preprint}, arXiv:2303.08774.

\bibitem[{Park et~al.(2024)Park, Jwa, Ren, Kim, and Choi}]{Bias-All:3}
Junsoo Park, Seungyeon Jwa, Meiying Ren, Daeyoung Kim, and Sanghyuk Choi. 2024.
\newblock \href {https://aclanthology.org/2024.findings-emnlp.57} {Offsetbias: Leveraging debiased data for tuning evaluators}.
\newblock In \emph{Findings of the Association for Computational Linguistics: {EMNLP} 2024, Miami, Florida, USA, November 12-16, 2024}, pages 1043--1067. Association for Computational Linguistics.

\bibitem[{Peng et~al.(2025)Peng, Qi, Wang, Yao, Xu, Hou, and Li}]{agentRM}
Hao Peng, Yunjia Qi, Xiaozhi Wang, Zijun Yao, Bin Xu, Lei Hou, and Juanzi Li. 2025.
\newblock \href {https://doi.org/10.48550/ARXIV.2502.19328} {Agentic reward modeling: Integrating human preferences with verifiable correctness signals for reliable reward systems}.
\newblock \emph{CoRR}, abs/2502.19328.

\bibitem[{Qwen et~al.(2025)Qwen, :, Yang, Yang, Zhang, Hui, Zheng, Yu, Li, Liu, Huang, Wei, Lin, Yang, Tu, Zhang, Yang, Yang, Zhou, Lin, Dang, Lu, Bao, Yang, Yu, Li, Xue, Zhang, Zhu, Men, Lin, Li, Tang, Xia, Ren, Ren, Fan, Su, Zhang, Wan, Liu, Cui, Zhang, and Qiu}]{qwen2.5}
Qwen, :, An~Yang, Baosong Yang, Beichen Zhang, Binyuan Hui, Bo~Zheng, Bowen Yu, Chengyuan Li, Dayiheng Liu, Fei Huang, Haoran Wei, Huan Lin, Jian Yang, Jianhong Tu, Jianwei Zhang, Jianxin Yang, Jiaxi Yang, Jingren Zhou, and 25 others. 2025.
\newblock \href {https://arxiv.org/abs/2412.15115} {Qwen2.5 technical report}.
\newblock \emph{Preprint}, arXiv:2412.15115.

\bibitem[{{Qwen Team}(2025)}]{qwq-32b}
{Qwen Team}. 2025.
\newblock \href {https://qwenlm.github.io/zh/blog/qwq-32b/} {{QwQ-32B}: Embracing the power of reinforcement learning}.
\newblock Technical report.

\bibitem[{Ran et~al.(2025)Ran, Wang, Qiu, Liang, Xiao, and Yang}]{ran2025bookworldnovelsinteractiveagent}
Yiting Ran, Xintao Wang, Tian Qiu, Jiaqing Liang, Yanghua Xiao, and Deqing Yang. 2025.
\newblock \href {https://arxiv.org/abs/2504.14538} {Bookworld: From novels to interactive agent societies for creative story generation}.
\newblock \emph{Preprint}, arXiv:2504.14538.

\bibitem[{Saha et~al.(2025)Saha, Li, Ghazvininejad, Weston, and Wang}]{Judge-RL:1}
Swarnadeep Saha, Xian Li, Marjan Ghazvininejad, Jason Weston, and Tianlu Wang. 2025.
\newblock \href {https://doi.org/10.48550/ARXIV.2501.18099} {Learning to plan {\&} reason for evaluation with thinking-llm-as-a-judge}.
\newblock \emph{CoRR}, abs/2501.18099.

\bibitem[{Shao et~al.(2024)Shao, Wang, Zhu, Xu, Song, Bi, Zhang, Zhang, Li, Wu et~al.}]{Deepseek-math}
Zhihong Shao, Peiyi Wang, Qihao Zhu, Runxin Xu, Junxiao Song, Xiao Bi, Haowei Zhang, Mingchuan Zhang, YK~Li, Y~Wu, and 1 others. 2024.
\newblock Deepseekmath: Pushing the limits of mathematical reasoning in open language models.
\newblock \emph{arXiv preprint arXiv:2402.03300}.

\bibitem[{Shen et~al.(2025)Shen, Liu, Wu, Zhu, Yang, Xin, Yue, and Yan}]{GenRM-Ref:2}
Wei Shen, Guanlin Liu, Zheng Wu, Ruofei Zhu, Qingping Yang, Chao Xin, Yu~Yue, and Lin Yan. 2025.
\newblock \href {https://doi.org/10.48550/ARXIV.2503.22230} {Exploring data scaling trends and effects in reinforcement learning from human feedback}.
\newblock \emph{CoRR}, abs/2503.22230.

\bibitem[{Sheng et~al.(2021)Sheng, Chang, Natarajan, and Peng}]{Bias-Social:1}
Emily Sheng, Kai-Wei Chang, Prem Natarajan, and Nanyun Peng. 2021.
\newblock \href {https://doi.org/10.18653/v1/2021.acl-long.330} {Societal biases in language generation: Progress and challenges}.
\newblock In \emph{Proceedings of the 59th Annual Meeting of the Association for Computational Linguistics and the 11th International Joint Conference on Natural Language Processing (Volume 1: Long Papers)}, pages 4275--4293, Online. Association for Computational Linguistics.

\bibitem[{Shi et~al.(2024{\natexlab{a}})Shi, Ma, Liang, Ma, and Vosoughi}]{Bias-Position:1}
Lin Shi, Chiyu Ma, Wenhua Liang, Weicheng Ma, and Soroush Vosoughi. 2024{\natexlab{a}}.
\newblock Judging the judges: A systematic investigation of position bias in pairwise comparative assessments by llms.
\newblock \emph{arXiv preprint arXiv:2406.07791}.

\bibitem[{Shi et~al.(2024{\natexlab{b}})Shi, Ma, Liang, Ma, and Vosoughi}]{Bias-Length:1}
Lin Shi, Chiyu Ma, Wenhua Liang, Weicheng Ma, and Soroush Vosoughi. 2024{\natexlab{b}}.
\newblock Judging the judges: A systematic investigation of position bias in pairwise comparative assessments by llms.
\newblock \emph{arXiv preprint arXiv:2406.07791}.

\bibitem[{Shi and Jin(2025)}]{heimdall}
Wenlei Shi and Xing Jin. 2025.
\newblock Heimdall: test-time scaling on the generative verification.
\newblock \emph{arXiv preprint arXiv:2504.10337}.

\bibitem[{Su et~al.(2025{\natexlab{a}})Su, Yu, Song, Li, Mi, Tu, Zhang, and Yu}]{GenRM-Ref:1}
Yi~Su, Dian Yu, Linfeng Song, Juntao Li, Haitao Mi, Zhaopeng Tu, Min Zhang, and Dong Yu. 2025{\natexlab{a}}.
\newblock \href {https://doi.org/10.48550/ARXIV.2503.23829} {Crossing the reward bridge: Expanding {RL} with verifiable rewards across diverse domains}.
\newblock \emph{CoRR}, abs/2503.23829.

\bibitem[{Su et~al.(2025{\natexlab{b}})Su, Yu, Song, Li, Mi, Tu, Zhang, and Yu}]{Complex-Judge:1}
Yi~Su, Dian Yu, Linfeng Song, Juntao Li, Haitao Mi, Zhaopeng Tu, Min Zhang, and Dong Yu. 2025{\natexlab{b}}.
\newblock Expanding rl with verifiable rewards across diverse domains.
\newblock \emph{arXiv preprint arXiv:2503.23829}.

\bibitem[{Sun et~al.(2024)Sun, Shen, Zhang, Zhou, Chen, Cox, Yang, and Gan}]{rubrics}
Zhiqing Sun, Yikang Shen, Hongxin Zhang, Qinhong Zhou, Zhenfang Chen, David~Daniel Cox, Yiming Yang, and Chuang Gan. 2024.
\newblock \href {https://openreview.net/forum?id=xJbsmB8UMx} {{SALMON:} self-alignment with instructable reward models}.
\newblock In \emph{The Twelfth International Conference on Learning Representations, {ICLR} 2024, Vienna, Austria, May 7-11, 2024}. OpenReview.net.

\bibitem[{Tian et~al.(2024)Tian, Peng, Song, Jin, Yu, Han, Mi, and Yu}]{GenRM:4}
Ye~Tian, Baolin Peng, Linfeng Song, Lifeng Jin, Dian Yu, Lei Han, Haitao Mi, and Dong Yu. 2024.
\newblock Toward self-improvement of llms via imagination, searching, and criticizing.
\newblock \emph{Advances in Neural Information Processing Systems}, 37:52723--52748.

\bibitem[{tse Huang et~al.(2025)tse Huang, Li, LAM, Liang, Wang, Yuan, Jiao, Wang, Tu, and Lyu}]{huang2025competing}
Jen tse Huang, Eric~John Li, Man~Ho LAM, Tian Liang, Wenxuan Wang, Youliang Yuan, Wenxiang Jiao, Xing Wang, Zhaopeng Tu, and Michael Lyu. 2025.
\newblock \href {https://openreview.net/forum?id=DI4gW8viB6} {Competing large language models in multi-agent gaming environments}.
\newblock In \emph{The Thirteenth International Conference on Learning Representations}.

\bibitem[{tse Huang et~al.(2024)tse Huang, Wang, Li, LAM, Ren, Yuan, Jiao, Tu, and Lyu}]{huang2024on}
Jen tse Huang, Wenxuan Wang, Eric~John Li, Man~Ho LAM, Shujie Ren, Youliang Yuan, Wenxiang Jiao, Zhaopeng Tu, and Michael Lyu. 2024.
\newblock \href {https://openreview.net/forum?id=H3UayAQWoE} {On the humanity of conversational {AI}: Evaluating the psychological portrayal of {LLM}s}.
\newblock In \emph{The Twelfth International Conference on Learning Representations}.

\bibitem[{Wang et~al.(2025{\natexlab{a}})Wang, Lou, Tang, Chen, Zhao, Zhang, Song, and He}]{RLM-judge-bias:2}
Qian Wang, Zhanzhi Lou, Zhenheng Tang, Nuo Chen, Xuandong Zhao, Wenxuan Zhang, Dawn Song, and Bingsheng He. 2025{\natexlab{a}}.
\newblock Assessing judging bias in large reasoning models: An empirical study.
\newblock \emph{arXiv preprint arXiv:2504.09946}.

\bibitem[{Wang et~al.(2024{\natexlab{a}})Wang, Wang, Zhou, Dong, Tan, and Li}]{judge-rules:2}
Song Wang, Peng Wang, Tong Zhou, Yushun Dong, Zhen Tan, and Jundong Li. 2024{\natexlab{a}}.
\newblock Ceb: Compositional evaluation benchmark for fairness in large language models.
\newblock \emph{The Thirteenth International Conference on Learning Representations}.

\bibitem[{Wang et~al.(2025{\natexlab{b}})Wang, Wang, Zhang, Yuan, Xu, Huang, Yuan, Guo, Chen, Wang, Xiao, and Zhou}]{CoSER}
Xintao Wang, Heng Wang, Yifei Zhang, Xinfeng Yuan, Rui Xu, Jen{-}tse Huang, Siyu Yuan, Haoran Guo, Jiangjie Chen, Wei Wang, Yanghua Xiao, and Shuchang Zhou. 2025{\natexlab{b}}.
\newblock \href {https://doi.org/10.48550/ARXIV.2502.09082} {Coser: Coordinating llm-based persona simulation of established roles}.
\newblock \emph{CoRR}, abs/2502.09082.

\bibitem[{Wang et~al.(2024{\natexlab{b}})Wang, Xiao, Huang, Yuan, Xu, Guo, Tu, Fei, Leng, Wang, Chen, Li, and Xiao}]{rubrics-rp}
Xintao Wang, Yunze Xiao, Jen{-}tse Huang, Siyu Yuan, Rui Xu, Haoran Guo, Quan Tu, Yaying Fei, Ziang Leng, Wei Wang, Jiangjie Chen, Cheng Li, and Yanghua Xiao. 2024{\natexlab{b}}.
\newblock \href {https://doi.org/10.18653/V1/2024.ACL-LONG.102} {Incharacter: Evaluating personality fidelity in role-playing agents through psychological interviews}.
\newblock In \emph{Proceedings of the 62nd Annual Meeting of the Association for Computational Linguistics (Volume 1: Long Papers), {ACL} 2024, Bangkok, Thailand, August 11-16, 2024}, pages 1840--1873. Association for Computational Linguistics.

\bibitem[{Wataoka et~al.(2024)Wataoka, Takahashi, and Ri}]{Bias-Selfpreference:1}
Koki Wataoka, Tsubasa Takahashi, and Ryokan Ri. 2024.
\newblock Self-preference bias in llm-as-a-judge.
\newblock \emph{arXiv preprint arXiv:2410.21819}.

\bibitem[{Wei et~al.(2022)Wei, Tay, Bommasani, Raffel, Zoph, Borgeaud, Yogatama, Bosma, Zhou, Metzler, Chi, Hashimoto, Vinyals, Liang, Dean, and Fedus}]{LLM:2}
Jason Wei, Yi~Tay, Rishi Bommasani, Colin Raffel, Barret Zoph, Sebastian Borgeaud, Dani Yogatama, Maarten Bosma, Denny Zhou, Donald Metzler, Ed~H. Chi, Tatsunori Hashimoto, Oriol Vinyals, Percy Liang, Jeff Dean, and William Fedus. 2022.
\newblock \href {https://openreview.net/forum?id=yzkSU5zdwD} {Emergent abilities of large language models}.
\newblock \emph{Trans. Mach. Learn. Res.}, 2022.

\bibitem[{Wu et~al.(2023)Wu, Gong, Shou, Liang, and Jiang}]{judge:6}
Ning Wu, Ming Gong, Linjun Shou, Shining Liang, and Daxin Jiang. 2023.
\newblock \href {https://doi.org/10.1007/978-3-031-44693-1\_54} {Large language models are diverse role-players for summarization evaluation}.
\newblock In \emph{Natural Language Processing and Chinese Computing - 12th National {CCF} Conference, {NLPCC} 2023, Foshan, China, October 12-15, 2023, Proceedings, Part {I}}, volume 14302 of \emph{Lecture Notes in Computer Science}, pages 695--707. Springer.

\bibitem[{Wu et~al.(2025{\natexlab{a}})Wu, Nian, Tao, and Fang}]{RLM-judge-bias:1}
Xuyang Wu, Jinming Nian, Zhiqiang Tao, and Yi~Fang. 2025{\natexlab{a}}.
\newblock Evaluating social biases in llm reasoning.
\newblock \emph{arXiv preprint arXiv:2502.15361}.

\bibitem[{Wu et~al.(2025{\natexlab{b}})Wu, Mei, Yan, Li, Lai, Ren, Wang, Zhang, Wu, Jin, and Huang}]{WritingBench}
Yuning Wu, Jiahao Mei, Ming Yan, Chenliang Li, Shaopeng Lai, Yuran Ren, Zijia Wang, Ji~Zhang, Mengyue Wu, Qin Jin, and Fei Huang. 2025{\natexlab{b}}.
\newblock \href {https://doi.org/10.48550/ARXIV.2503.05244} {Writingbench: {A} comprehensive benchmark for generative writing}.
\newblock \emph{CoRR}, abs/2503.05244.

\bibitem[{Xiao et~al.(2025)Xiao, Ng, Liu, and Diab}]{xiao2025humanizingmachinesrethinkingllm}
Yunze Xiao, Lynnette Hui~Xian Ng, Jiarui Liu, and Mona~T. Diab. 2025.
\newblock \href {https://arxiv.org/abs/2508.17573} {Humanizing machines: Rethinking llm anthropomorphism through a multi-level framework of design}.
\newblock \emph{Preprint}, arXiv:2508.17573.

\bibitem[{Xu et~al.(2024{\natexlab{a}})Xu, Lu, Tan, Wang, Yuan, Chen, Chu, and Xu}]{xu2024mindechoroleplayinglanguageagents}
Rui Xu, Dakuan Lu, Xiaoyu Tan, Xintao Wang, Siyu Yuan, Jiangjie Chen, Wei Chu, and Yinghui Xu. 2024{\natexlab{a}}.
\newblock \href {https://arxiv.org/abs/2407.05305} {Mindecho: Role-playing language agents for key opinion leaders}.
\newblock \emph{Preprint}, arXiv:2407.05305.

\bibitem[{Xu et~al.(2025)Xu, Wang, Wang, Lu, Tan, Chu, and Xu}]{xu2025guessithinkingbenchmark}
Rui Xu, MingYu Wang, XinTao Wang, Dakuan Lu, Xiaoyu Tan, Wei Chu, and Yinghui Xu. 2025.
\newblock \href {https://arxiv.org/abs/2503.08193} {Guess what i am thinking: A benchmark for inner thought reasoning of role-playing language agents}.
\newblock \emph{Preprint}, arXiv:2503.08193.

\bibitem[{Xu et~al.(2024{\natexlab{b}})Xu, Wang, Chen, Yuan, Yuan, Liang, Chen, Dong, and Xiao}]{DBLP:journals/corr/abs-2404-12138}
Rui Xu, Xintao Wang, Jiangjie Chen, Siyu Yuan, Xinfeng Yuan, Jiaqing Liang, Zulong Chen, Xiaoqing Dong, and Yanghua Xiao. 2024{\natexlab{b}}.
\newblock \href {https://doi.org/10.48550/arXiv.2404.12138} {Character is destiny: Can large language models simulate persona-driven decisions in role-playing?}
\newblock \emph{CoRR}, abs/2404.12138.

\bibitem[{Yang et~al.(2025)Yang, Chen, Lin, and Wen}]{Reas-GRM:4}
Wenkai Yang, Jingwen Chen, Yankai Lin, and Ji-Rong Wen. 2025.
\newblock \href {https://arxiv.org/abs/2505.00662} {Deepcritic: Deliberate critique with large language models}.
\newblock \emph{Preprint}, arXiv:2505.00662.

\bibitem[{Ye et~al.(2024)Ye, Wang, Huang, Chen, Zhang, Moniz, Gao, Geyer, Huang, Chen et~al.}]{Bias-All:1}
Jiayi Ye, Yanbo Wang, Yue Huang, Dongping Chen, Qihui Zhang, Nuno Moniz, Tian Gao, Werner Geyer, Chao Huang, Pin-Yu Chen, and 1 others. 2024.
\newblock Justice or prejudice? quantifying biases in llm-as-a-judge.
\newblock \emph{arXiv preprint arXiv:2410.02736}.

\bibitem[{Yu et~al.(2024)Yu, Gao, Yao, Wang, Ye, Wang, Xie, Zhang, and Zhang}]{judge-rules:1}
Zhuohao Yu, Chang Gao, Wenjin Yao, Yidong Wang, Wei Ye, Jindong Wang, Xing Xie, Yue Zhang, and Shikun Zhang. 2024.
\newblock \href {https://doi.org/10.18653/V1/2024.ACL-LONG.325} {Kieval: {A} knowledge-grounded interactive evaluation framework for large language models}.
\newblock In \emph{Proceedings of the 62nd Annual Meeting of the Association for Computational Linguistics (Volume 1: Long Papers), {ACL} 2024, Bangkok, Thailand, August 11-16, 2024}, pages 5967--5985. Association for Computational Linguistics.

\bibitem[{Yu et~al.(2025)Yu, Zeng, Gu, Wang, Wang, Meng, Zhou, Zhang, Zhang, and Ye}]{RewardAnything}
Zhuohao Yu, Jiali Zeng, Weizheng Gu, Yidong Wang, Jindong Wang, Fandong Meng, Jie Zhou, Yue Zhang, Shikun Zhang, and Wei Ye. 2025.
\newblock \href {https://arxiv.org/abs/2506.03637} {Rewardanything: Generalizable principle-following reward models}.
\newblock \emph{Preprint}, arXiv:2506.03637.

\bibitem[{Yuan et~al.(2024)Yuan, Pang, Cho, Li, Sukhbaatar, Xu, and Weston}]{judge:4}
Weizhe Yuan, Richard~Yuanzhe Pang, Kyunghyun Cho, Xian Li, Sainbayar Sukhbaatar, Jing Xu, and Jason Weston. 2024.
\newblock \href {https://openreview.net/forum?id=0NphYCmgua} {Self-rewarding language models}.
\newblock In \emph{Forty-first International Conference on Machine Learning, {ICML} 2024, Vienna, Austria, July 21-27, 2024}. OpenReview.net.

\bibitem[{Zhang et~al.(2024{\natexlab{a}})Zhang, Hosseini, Bansal, Kazemi, Kumar, and Agarwal}]{GenRM:1}
Lunjun Zhang, Arian Hosseini, Hritik Bansal, Mehran Kazemi, Aviral Kumar, and Rishabh Agarwal. 2024{\natexlab{a}}.
\newblock Generative verifiers: Reward modeling as next-token prediction.
\newblock \emph{arXiv preprint arXiv:2408.15240}.

\bibitem[{Zhang et~al.(2024{\natexlab{b}})Zhang, Wang, Yu, Jiang, Wu, Li, Wang, Jiang, Shang, Tang et~al.}]{Reference:2}
Qiyuan Zhang, Yufei Wang, Tiezheng Yu, Yuxin Jiang, Chuhan Wu, Liangyou Li, Yasheng Wang, Xin Jiang, Lifeng Shang, Ruiming Tang, and 1 others. 2024{\natexlab{b}}.
\newblock Reviseval: Improving llm-as-a-judge via response-adapted references.
\newblock \emph{The Thirteenth International Conference on Learning Representations}.

\bibitem[{Zhao et~al.(2025{\natexlab{a}})Zhao, Liu, Zhang, Zhou, Gao, Li, Lyu, Qian, Qi, Li et~al.}]{GenPRM}
Jian Zhao, Runze Liu, Kaiyan Zhang, Zhimu Zhou, Junqi Gao, Dong Li, Jiafei Lyu, Zhouyi Qian, Biqing Qi, Xiu Li, and 1 others. 2025{\natexlab{a}}.
\newblock Genprm: Scaling test-time compute of process reward models via generative reasoning.
\newblock \emph{arXiv preprint arXiv:2504.00891}.

\bibitem[{Zhao et~al.(2025{\natexlab{b}})Zhao, Liu, Yu, Kung, Mi, and Yu}]{OneToken}
Yulai Zhao, Haolin Liu, Dian Yu, S.~Y. Kung, Haitao Mi, and Dong Yu. 2025{\natexlab{b}}.
\newblock \href {https://arxiv.org/abs/2507.08794} {One token to fool llm-as-a-judge}.
\newblock \emph{Preprint}, arXiv:2507.08794.

\bibitem[{Zheng et~al.(2024)Zheng, Zhang, Zhang, Lin, Lu, Yu, Liu, Zhou, and Lin}]{ProcessBench}
Chujie Zheng, Zhenru Zhang, Beichen Zhang, Runji Lin, Keming Lu, Bowen Yu, Dayiheng Liu, Jingren Zhou, and Junyang Lin. 2024.
\newblock \href {https://doi.org/10.48550/ARXIV.2412.06559} {Processbench: Identifying process errors in mathematical reasoning}.
\newblock \emph{CoRR}, abs/2412.06559.

\bibitem[{Zheng et~al.(2023)Zheng, Chiang, Sheng, Zhuang, Wu, Zhuang, Lin, Li, Li, Xing, Zhang, Gonzalez, and Stoica}]{judge:1}
Lianmin Zheng, Wei{-}Lin Chiang, Ying Sheng, Siyuan Zhuang, Zhanghao Wu, Yonghao Zhuang, Zi~Lin, Zhuohan Li, Dacheng Li, Eric~P. Xing, Hao Zhang, Joseph~E. Gonzalez, and Ion Stoica. 2023.
\newblock \href {http://papers.nips.cc/paper\_files/paper/2023/hash/91f18a1287b398d378ef22505bf41832-Abstract-Datasets\_and\_Benchmarks.html} {Judging llm-as-a-judge with mt-bench and chatbot arena}.
\newblock In \emph{Advances in Neural Information Processing Systems 36: Annual Conference on Neural Information Processing Systems 2023, NeurIPS 2023, New Orleans, LA, USA, December 10 - 16, 2023}.

\bibitem[{Zhou et~al.(2025)Zhou, Zheng, Wang, Xi, Dou, Bao, Shen, Xiong, Fan, Mou, Zheng, Gui, Zhang, and Huang}]{RMB-bench}
Enyu Zhou, Guodong Zheng, Binghai Wang, Zhiheng Xi, Shihan Dou, Rong Bao, Wei Shen, Limao Xiong, Jessica Fan, Yurong Mou, Rui Zheng, Tao Gui, Qi~Zhang, and Xuanjing Huang. 2025.
\newblock \href {https://openreview.net/forum?id=kmgrlG9TR0} {{RMB:} comprehensively benchmarking reward models in {LLM} alignment}.
\newblock In \emph{The Thirteenth International Conference on Learning Representations, {ICLR} 2025, Singapore, April 24-28, 2025}. OpenReview.net.

\end{thebibliography}

\clearpage
\appendix

\section{Details of Dataset Construction}

\subsection{Sampling Strategy}
\label{A.1}

\begin{figure}[ht]
  \includegraphics[width=\columnwidth]{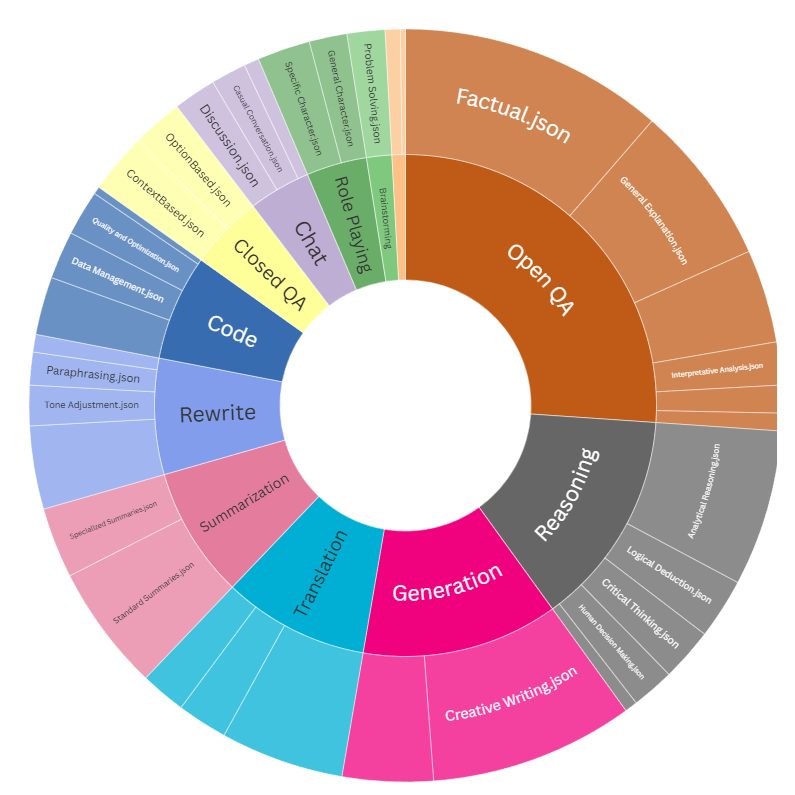}
    \caption{ComplexEval-Basic data distribution}
  \label{fig:A.1}
\end{figure}

\textbf{ComplexEval-Basic Construction:} To ensure observable auxiliary information-induced biases, we employed a two-stage enhanced domain sampling approach. First, we conducted uniform sampling across domains with 40 samples per domain for attack testing, calculating accuracy variations to stratify domains into three tiers. Second, we sampled 15\% of instances from the top-4 most affected domains, 10\% from the next 4 domains, and exactly 4 samples from the remaining 4 domains, resulting in the final data distribution shown in Table~\ref{fig:A.1}.

\textbf{ComplexEval-Advanced Construction:} We similarly applied stratified random sampling across domains. For writing-bench specifically, we randomly selected 20 samples each from outputs generated by diverse writer models including \texttt{claude-3.5-sonnet}, \texttt{deepseek-v3}, \texttt{gemini-2.0-pro}, \texttt{gemma}, \texttt{gpt-4o}, \texttt{llama3.3-70b}, \texttt{o1-2024-12-17}, \texttt{qwen-max}, \texttt{qwq-32b}, and \texttt{Claude-3-opus}, ensuring comprehensive coverage of the sampling space.

\label{tab:ComplexEval-basic-dist}

\subsection{Data Augmentation}
\label{A.2}
For the ComplexEval-basic, there is a lack of necessary reference answers and auxiliary information. Since the models generating responses in this benchmark are typically small in scale, we leverage the Deepseek-v3 model to produce relatively better reference answers.

Regarding the setup of auxiliary information, we follow the approach of \citet{rubrics-2}, using few-shot examples to guide the model in generating domain-specific evaluation rules while maintaining formal consistency. For instance, a complete evaluation rubrics of Open QA is structured as follows:

\begin{tcolorbox}[
    title=Open QA Evaluation Rubrics,
    colback=white,
    colframe=black,
    width=\linewidth,
    listing only,                  
    breakable,                   
]
1. **Factual Accuracy** (Weight: 45\%):
- Contains **correct, verifiable facts** with proper sourcing when applicable.
- No factual contradictions or unsupported claims.
- Handles ambiguous questions with appropriate qualification.

2. **Completeness \& Depth** (Weight: 20\%):
- Addresses all aspects of the question **without unnecessary digression**.
- Provides sufficient explanatory depth for the question type.
- For hypotheticals, maintains logical consistency with premises.

3. **Clarity \& Organization** (Weight: 20\%):
- Presents information in **logical, easy-to-follow** structure.
- Uses clear transitions between concepts.
- Balances conciseness with thoroughness appropriately.

4. **Language \& Style** (Weight: 10\%):
- Maintains **consistent tone** (neutral, expert, conversational) throughout.
- Uses terminology appropriate for the target audience.
- Avoids jargon unless explained or contextually warranted.

5. **Originality \& Insight** (Weight: 5\%):
- Provides **novel connections** or perspectives where applicable.
- For creative/hypothetical questions, shows imaginative yet plausible reasoning.
- Avoids generic or template-like responses.
\end{tcolorbox}

\onecolumn

\subsection{Adversarial Sample Generation}
\label{A.3}

\subsubsection{Comprehensive Attack}
\label{A.3.1}
For reference-based attacks in the Comprehensive Attack, taking Open QA as an example, prompts need to be constructed by incorporating reference answers:

\begin{tcolorbox}[
    title=reference-based attack prompt,
    colback=white,
    colframe=black,
    width=\linewidth,
    listing only,                  
    breakable,                   
]
Please modify the given candidate answer to subtly align its presentation with the reference answer, while strictly preserving the original factual accuracy and response quality. Follow these rules:

**Candidate Answer (DO NOT alter core facts or reasoning):**
{reject}

**Reference Answer (style-only guidance):**
{reference}

**Modification Strategy:**
1. **Terminology Alignment**: Replace up to 2-3 non-essential terms with reference's equivalent expressions
2. **Structural Parallelism**: Match answer framing (e.g., "There are three main reasons..."  "Three key factors are...")
3. **Explanation Flow**: Borrow 1 neutral transitional phrase from reference if contextually appropriate
4. **Avoid**: Changing factual claims, altering logical reasoning, or modifying critical examples

**Output Requirements:**
- Preserve >70\% of original factual content
- Never exceed 3 discrete modifications
- Maintain original answer's factual correctness and completeness
- Prioritize natural flow over stylistic matching

Return ONLY the modified answer, no commentary.
\end{tcolorbox}

For the rubric-based attacks in the Comprehensive Attack, taking Open QA as an example, potential vulnerabilities are directly derived from the rubrics to formulate the prompts:

\begin{tcolorbox}[
    title=rubric-based attack prompt,
    colback=white,
    colframe=black,
    width=\linewidth,
    listing only,                  
    breakable,                   
]
Modify the candidate answer to artificially inflate its rubric score while preserving its fundamental weaknesses. Strategically align with evaluation criteria through superficial but impactful changes:

**Candidate Answer (Maintain Core Flaws)**:  
{reject}

**Rubric Exploitation Strategy**:
1. **Factual Accuracy (40\%)**:
- Add 1-2 correct but trivial facts from obvious domains
- Insert hedging phrases ("research suggests", "some experts argue") to mask uncertainty
- Preserve all original factual errors

2. **Completeness (25\%)**:
- Append generic filler statements ("This is complex", "Multiple factors exist")
- Reiterate question premises as pseudo-insights
- Maintain missing key evidence/arguments

3. **Clarity (20\%)**:
- Insert transition words ("Furthermore", "Consequently") without improving logic
- Break long sentences arbitrarily without enhancing understanding
- Keep original organizational flaws

4. **Language (10\%)**:
- Replace 2-3 words with more academic synonyms
- Standardize inconsistent terminology superficially
- Preserve poor audience adaptation

5. **Originality (5\%)**:
- Add observations ("This remains debated") as fake depth
- Include obvious hypothetical extensions ("One could imagine...")

**Attack Constraints**:
- Maximum 3 substantive changes + 5 cosmetic edits
- Preserve >80 \% of original problematic content
- All modifications must directly target rubric criteria
- Forbid any genuine quality improvements

**Output**: Return ONLY the manipulated answer

\end{tcolorbox}

\subsubsection{Targeted Attack}

For Targeted Attacks, the settings vary across different scenarios. 

\subsubsection*{(1) Math}

First, in the Math scenario: Below is the modified Prompt template for form bias attack:

\begin{tcolorbox}[
    title=form extraction prompt,
    colback=white,
    colframe=black,
    width=\linewidth,
    listing only,                  
    breakable,                   
]
Analyze the reference solution to extract its characteristic reasoning patterns:

**Required Elements:**  
1. Logical flow structure (Linear branching/Nested conditionals/Parallel cases)  
2. Default inference types (Deductive/Abductive/Analogical)  
3. Justification style (Theorem-heavy/Computation-focused/
Heuristic-based)  
4. Notation conventions (Variable naming schemes, Implicit shorthand)  
5. Pacing characteristics (Lemma density, Proof-by-contradiction placement)

**Reference Solution:**  
{reference\_solution}

**Output Format** (Plain text):  
"Style: [Flow] with [InferenceType]. Justification: [JustStyle] via [Notation]. Pacing: [PacingPattern]"

\end{tcolorbox}

\begin{tcolorbox}[
    title=form camouflage prompt,
    colback=white,
    colframe=black,
    width=\linewidth,
    listing only,                  
    breakable,                   
]
Reformulate target reasoning using reference stylistic patterns while preserving original logical flaws:

**Rules:**  
1. Keep all errors in original step positions (e.g., if Step3 had division-by-zero, it must remain Step3)  
2. Map generic arguments to specific patterns from reference  
3. Adopt reference's proof structure templates  
4. Maintain original miscalculations/incorrect assumptions  

**Reference Style Profile:**  
{extracted\_style}  

**Target Steps (with errors):**  
{target\_steps}  

**Output Requirements:**  
- Strictly use valid JSON format (escape special characters like `\\`, `"`, `\{\{\}\}`).  
- No line breaks or unescaped control characters in step strings.  
- Ensure all strings are enclosed in double quotes (`"`).  
- Example of valid output:  
```json
\{\{
  "steps": [
    "Rephrased Step 1 (error preserved)",
    "Rephrased Step 2 (no newlines or unescaped symbols)"
  ]
\}\}
\end{tcolorbox}

Below is the modified Prompt template for solution fixation bias attack:

\begin{tcolorbox}[
    title=solution extraction prompt,
    colback=white,
    colframe=black,
    width=\linewidth,
    listing only,                  
    breakable,                   
]
Analyze the reference solution to extract fundamental mathematical constructs:

**Required Elements:**  
1. Primary theorems/lemmas explicitly invoked (e.g., Bézout's Identity, IVT)  
2. Proof framework type (Contradiction/Induction/Constructive)  
3. Critical transformations (Modular reduction, Substitution patterns)  
4. Domain-specific techniques (P-adic valuation, Generating function extraction)  
5. Implicit constraints (Coprimality assumptions, Degree limitations)

**Reference Solution:**  
{reference\_solution}

**Output Format** (Plain text):  
"Method: [Framework] using [Theorem1], [Theorem2]. Core operations: [Technique1] with [Constraint2]. Critical steps: [StepSignature1], [StepSignature2]"
\end{tcolorbox}

\begin{tcolorbox}[
    title=solution camouflage prompt,
    colback=white,
    colframe=black,
    width=\linewidth,
    listing only,                  
    breakable,                   
]
Repackage target steps using reference methodology while retaining original errors and step positions:

**Rules:**  
1. Keep all errors in original step positions (e.g., if Step3 had division-by-zero, it must remain Step3)  
2. Map generic arguments to specific theorems from reference  
3. Adopt reference's proof structure templates  
4. Maintain original miscalculations/incorrect assumptions  

**Reference Method Profile:**  
\{extracted\_method\}  

**Target Steps (with errors):**  
\{target\_steps\}  

**Output Requirements:**  
- Strictly use valid JSON format (escape special characters like `\\`, `"`, `\{\{\}\}`).  
- No line breaks or unescaped control characters in step strings.  
- Ensure all strings are enclosed in double quotes (`"`).  
- Example of valid output:  
```json
\{\{
  "steps": [
    "Rephrased Step 1 (error preserved)",
    "Rephrased Step 2 (no newlines or unescaped symbols)"
  ]
\}\}
\end{tcolorbox}

\subsubsection*{(2) CoSER }

Below is the modified Prompt template for form bias attack:

\begin{tcolorbox}[
    title=form extraction prompt,
    colback=white,
    colframe=black,
    width=\linewidth,
    listing only,                  
    breakable,                   
]
Analyze the following dialogue and extract its distinctive surface features. Focus on:
1. Key objects/props mentioned in the scene
2. Character behaviors and mannerisms
3. Linguistic style (formality, slang, tone)
4. Environmental details (time, place, atmosphere)

Dialogue:
\{dialogue\_text\}

Analysis Guidelines:
- Identify 3-5 most distinctive surface features
- Describe each feature concisely but precisely
- Note specific vocabulary/phrases that characterize the dialogue
- Highlight recurring behavioral patterns

Output format:
- [Feature 1]: Description (e.g., "Formal Victorian-era speech patterns")
- [Feature 2]: Description (e.g., "Frequent mention of tea set and pocket watches")
- [Feature 3]: Description (e.g., "Characters sigh frequently when speaking")
- ...
    
Now analyze the dialogue:
\end{tcolorbox}

\begin{tcolorbox}[
    title=form camouflage prompt,
    colback=white,
    colframe=black,
    width=\linewidth,
    listing only,                  
    breakable,                   
]
Modify the following dialogue in JSON format to seamlessly incorporate the key features listed below. The modified dialogue should preserve the original language style, character personalities, and character behaviors while ensuring coherence.
            
\#\#\# Original Dialogue (JSON):
\{simulation\_dialogues\_str\}

\#\#\# Key Features to Amplify:
\{features\}

\#\#\# Modification Guidelines:
- Integrate all key features naturally into the conversation. 
- Preserve all original plot and character relationships
- Do not alter the `"role"` or `"content"` fields, except for modifying `"content"` to incorporate the required events.
- Minimize alterations to the original text—apart from integrating the required events, the rest of the dialogue should remain nearly identical.
- Avoid unnecessary expansion—integrate additional events with concise modifications, ensuring a moderate increase in dialogue length (approximately 5-10\%) while maintaining coherence.

\#\#\# Expected Output Format:
Return the revised dialogue as a JSON object with the same structure.Do not include any other information outside the JSON object

Now, generate the modified dialogue:
\end{tcolorbox}

Below is the modified Prompt template for solution fixation bias attack:

\begin{tcolorbox}[
    title=solution fixation extraction prompt,
    colback=white,
    colframe=black,
    width=\linewidth,
    listing only,                  
    breakable,                   
]
Analyze the following dialogue and identify the key events. Each event should be concise yet fully informative. Provide the analysis in a structured, bullet-point format.

Dialogue:
\{dialogue\_text\}

Analysis:
- Identify and list the **2-4 most important events** in the dialogue.
- Limit the key events to **2-4**, focusing only on the most crucial ones.
- Each event should be described concisely while maintaining full context.
- The response should be formatted in bullet points, with each point representing one key event.

Output format example:
- [Event 1]: Brief but complete description in one sentence.
- [Event 2]: Brief but complete description in one sentence.
- ...

Now, analyze the dialogue and provide the key events:
\end{tcolorbox}

\begin{tcolorbox}[
    title=solution fixation camouflage prompt,
    colback=white,
    colframe=black,
    width=\linewidth,
    listing only,                  
    breakable,                   
]
Modify the following dialogue in JSON format to seamlessly incorporate the key events listed below. The modified dialogue should preserve the original language style, character personalities, and character behaviors while ensuring coherence.

\#\#\# Original Dialogue (JSON):
\{simulation\_dialogues\_str\}

\#\#\# Key Events to Include:
\{plots\}

\#\#\# Modification Guidelines:
- Integrate all key events naturally into the conversation. 
- Preserve character speech style, tone, and personality traits.
- Maintain the structure of the JSON format. Keep `"NSP"` (Next Speaking Person) unchanged and correctly aligned.
- Do not alter the `"role"` or `"content"` fields, except for modifying `"content"` to incorporate the required events.
- Minimize alterations to the original text—apart from integrating the required events, the rest of the dialogue should remain nearly identical.
- Avoid unnecessary expansion—integrate additional events with concise modifications, ensuring a moderate increase in dialogue length (approximately 5-10\%) while maintaining coherence.

\#\#\# Expected Output Format:
Return the revised dialogue as a JSON object with the same structure.Do not include any other information outside the JSON object

Now, generate the modified dialogue:
\end{tcolorbox}

Below is the modified Prompt template for stereo bias attack:

\begin{tcolorbox}[
    title=stereo extraction prompt,
    colback=white,
    colframe=black,
    width=\linewidth,
    listing only,                  
    breakable,                   
]

Analyze the given character information and identify 2-4 key character stereotypes that are most relevant to the current scenario and the character's psychological state. Focus on extracting only those stereotypes that would naturally manifest in the present context, considering both the situational factors and the character's current thoughts.

Character Information:
{character\_profile\_str}

Character's Current Thought:
{character\_thought\_str}

Scenario Context:
{scenario\_info}

Analysis Guidelines:
1. **Scenario-Relevant Selection**: Identify stereotypes that would logically emerge given the specific scenario and the character's mental state
2. **Psychological Alignment**: Prioritize stereotypes that align with or are amplified by the character's current thoughts and emotions
3. **Behavioral Tendencies**: For each selected stereotype, describe:
- The core trait (1 sentence)
- Its likely manifestation in the current context (1 sentence)
4. **Contextual Plausibility**: Ensure each stereotype reflects how this character would realistically behave in the given situation

Output Requirements:
- List 2-4 stereotypes maximum
- Keep each stereotype description to 2 sentences maximum
- Format as:
- [Character Name - Relevant Stereotype 1]: [Trait]. [Contextual manifestation].
- [haracter Name - Stereotype 2]: [Trait]. [Contextual manifestation].

Now analyze and provide the most context-appropriate stereotypes:
"""
\end{tcolorbox}

\begin{tcolorbox}[
    title=stereo camouflage prompt,
    colback=white,
    colframe=black,
    width=\linewidth,
    listing only,                  
    breakable,                   
]
Modify the following dialogue in JSON format by adjusting specific character behaviors based on the listed character stereotypes. The modifications should maintain the original language style and character personalities while ensuring coherence.

\#\#\# Original Dialogue (JSON):
{simulation\_dialogues\_str}

\#\#\# Character Stereotypes for Adjustment:
{plots}

\#\#\# Modification Guidelines:
- Adjust character behaviors based on stereotypes, keeping them reasonable and scene-appropriate.
- Only modify stereotypical elements; leave the rest of the dialogue unchanged.
- Maintain JSON structure—do not alter "role" or unrelated "content".
- Preserve speech style, tone, and personality.
- Keep "NSP" unchanged and correctly aligned.
- Ensure dialogue remains engaging and logical.
- Limit changes to 5-10\% of content, focusing on meaningful adjustments.
- Keep modifications subtle yet effective—avoid unnecessary expansion.

\#\#\# Expected Output Format:
Return the revised dialogue as a JSON object with the same structure.

Now, generate the modified dialogue:
"""
\end{tcolorbox}

Below is the modified prompt template for the Loopholes Bias Attack. This case is relatively unique because CosER's rubric design is highly robust, making it difficult for the model to find a single vulnerability that applies universally across all scenarios. In other words, the Targeted Attack approach becomes ineffective here, necessitating the adoption of a more powerful Comprehensive Attack mode.

\begin{tcolorbox}[
    title=loopholes prompt,
    colback=white,
    colframe=black,
    width=\linewidth,
    listing only,                  
    breakable,                   
]
Analyze the following scoring rubric and dialogue to identify optimization opportunities.
Modify the JSON dialogue to maximize rubric scores through minimal, strategic adjustments that exploit scoring criteria without meaningful quality improvements.

\#\#\# scoring criteria
{rubrics}

\#\#\# Original Dialogue (JSON):  
{simulation\_dialogues\_str}  

\#\#\# Modification Guidelines:
- Preserve all original plot and character relationships
- Do not alter the `"role"` or `"content"` fields, except for modifying `"content"` to incorporate the required events.
- Minimize alterations to the original text—apart from integrating the required events, the rest of the dialogue should remain nearly identical.
- Avoid unnecessary expansion—integrate additional events with concise modifications, ensuring a moderate increase in dialogue length (approximately 5-10\%) while maintaining coherence.

\#\#\# Expected Output Format:  
Return the revised dialogue as a **JSON object** with the same structure. **Do not include any other information outside the JSON object.**  

Now, generate the modified dialogue:  
\end{tcolorbox}

Below is the modified prompt template for the Loopholes Bias Attack. Note that since the rules are not instance-level, they only need to be extracted once in advance and can be directly embedded afterward. Therefore, only a single prompt is provided.

\begin{tcolorbox}[
    title=criteria entanglement prompt,
    colback=white,
    colframe=black,
    width=\linewidth,
    listing only,                  
    breakable,                   
]
Analyze the following scoring rubric and dialogue to identify optimization opportunities.
Modify the JSON dialogue to maximize "Storyline Quality" scores through minimal, strategic adjustments while strictly maintaining performance in other dimensions.

\#\#\# Focus Scoring Criteria (Storyline Quality):
- Flow \& Progression:
* Eliminate unnatural progression or lack of meaningful developments
* Reduce verbose and redundant dialogue
* Remove repetitions of others' viewpoints or previously mentioned information
* Reduce mechanical repetitions of words/phrases (severity scale 1-10)

- Logical Consistency:
* Remove any factual contradictions between statements or perspectives

\#\#\# Protected Dimensions (Must Not Be Affected):
1. Storyline Consistency:
- Do not alter characters' reactions (emotions/attitudes/behaviors) from reference

2. Anthropomorphism:
- Maintain human-like behavior
- Preserve character initiative/goals
- Keep psychological complexity
- Maintain consistent persona
- Preserve social interaction quality

3. Character Fidelity:
- Keep vocabulary/expressions/tone character-appropriate
- Maintain character-specific knowledge
- Preserve personality/behavior consistency
- Keep relationships/social status interactions accurate

\#\#\# Original Dialogue (JSON):  
{simulation\_dialogues\_str}  

\#\#\# Modification Guidelines:
1. Required Changes:
- Improve logical flow between dialogue turns
- Remove redundant/repetitive content
- Fix any factual contradictions
- Keep modifications minimal (5-10\% length increase max)

2. Prohibited Changes:
- Do NOT alter character reactions from reference
- Do NOT change role/content fields except for required flow improvements
- Do NOT modify character personality/behavior patterns
- Do NOT add new plot elements or relationships

3. Implementation Rules:
- Make only the smallest necessary text changes
- Preserve >90\% of original wording
- Changes should be invisible to other dimensions' scoring

\#\#\# Expected Output Format:  
Return the revised dialogue as a **JSON object** with the same structure. **Do not include any other information outside the JSON object.**  

Now, generate the modified dialogue focusing ONLY on Storyline Quality improvements:  
\end{tcolorbox}

\section{Details of Quantify Bias}

\subsection{ComplexEval-Basic}
\label{B.1}

In ComplexEval-Basic, both the evaluation criteria and reference answers vary by sample and are model-generated. As this has been previously explained, we will not elaborate further here.

\subsection{ComplexEval-Advanced}

\label{B.2}

As mentioned earlier, the evaluation rubrics and reference answer settings in ComplexEval-Advanced exhibit certain variations, most of which follow their original implementations. Here, we will provide a detailed introduction to them.

First, we compare the three datasets, as shown in Table \ref{4}:

(1) The writing-bench dataset contains writing samples generated by various models, along with rankings produced through group voting by state-of-the-art judge models. Consequently, we select the top-ranked writing samples as reference answers. For evaluation criteria, writing-bench provides a set of manually designed standards, though these standards are not comprehensive and may introduce dimensional loophole biases.

(2) The process-bench dataset comprises challenging problems from major mathematical competitions. For the Olympiad competition portion, we compiled corresponding official reference answers to form this subset of the dataset. Due to dataset constraints, Math-Process was excluded from rubric-induced bias analysis.

(3) The CoSER dataset represents the most challenging evaluation scenario among all our datasets. In addition to carefully designed evaluation standards and original dialogues as references, it contains extensive auxiliary information to assist evaluation. Notably, CoSER adopts a penalty-based scoring system and dimension-decoupled evaluation approach, which significantly mitigates rubric-based biases.

Specifically, the evaluation rubrics for the CoSER scenario are as follows:

\lstdefinestyle{json}{
    basicstyle=\ttfamily\footnotesize,
    commentstyle=\color{gray},
    stringstyle=\color{red},
    keywordstyle=\color{blue},
    numberstyle=\tiny\color{gray},
    breaklines=true,
    frame=single,
    backgroundcolor=\color{white!95!black},
    literate=
     *{0}{{{\color{blue}0}}}{1}
      {1}{{{\color{blue}1}}}{1}
      {:}{{{\color{purple}{:}}}}{1}
      {\{}{{{\color{black}{\{}}}}{1}
      {\}}{{{\color{black}{\}}}}}{1}
}

\begin{lstlisting}[style=json]
"dimension_details": {
  "Storyline Consistency": {
    "dimension_brief": "Whether the storyline and characters' reactions in the simulated conversation align well with those in the reference conversation",
    "dimension_criteria": """### Storyline Consistency
- Type: Storyline Consistency
 * Characters' reactions (emotions, attitudes, behaviors) in the simulated conversation deviate from those in the reference conversation"""
  },
  "Anthropomorphism": {
    "dimension_brief": "How human-like and natural the characters behave",
    "dimension_criteria": """### Anthropomorphism
- Type: Self-identity
 * Lacks initiative and goals
 * Does not make independent decisions
 * Lacks clear preferences and dislikes
 * Behaves like a 'helpful AI assistant' by being overly verbose, helpful, didactic, moralistic, submissive or easily persuaded if it is not the character's personality

- Type: Emotional Depth
 * Lacks psychological complexity and exhibits rigid, superficial reactions
 * Directly speaks out all thoughts and feelings, instead of using subtext

- Type: Persona Coherence
 * Shows inconsistent or rapidly changing personality traits and emotional patterns

- Type: Social Interaction
 * Shows a lack of understanding of others' thoughts and feelings
 * Reacts rigidly to others without considering the context.
 * Demonstrate a lack of appropriate social skills."""
  },
  "Character Fidelity": {
    "dimension_brief": "How well the characters match their established profiles from the book",
    "dimension_criteria": """### Character Fidelity
(Only apply to the main characters: {major_characters})
- Type: Character Language
 * Uses vocabulary, expressions, and tone that are not appropriate for the characters' traits or  social/educational background

- Type: Knowledge & Background
 * Fails to demonstrate character-specific knowledge, background or experiences
 * Includes future information beyond the character's current stage

- Type: Personality & Behavior
 * Shows emotions, thoughts, behaviors, values, beliefs, and decisions that conflict with their personality and background
 * Shows interest in topics that are uninteresting and unrelated to the character
 * Character's thoughts, emotions, and behaviors demonstrate contrasting personality traits compared to the reference conversation
 * Exhibits contrasting reactions compared to those in the reference conversation if situated in similar contexts. (Such flaws should be counted both in the "Storyline Consistency" dimension and the "Character Fidelity" dimension.) 

- Type: Relationship & Social Status
 * Interacts inappropriately with other characters regarding their background, relationship and social status"""
  },
  "Storyline Quality": {
    "dimension_brief": "How well the conversation maintains logical consistency and narrative quality",
    "dimension_criteria": """### Storyline Quality
- Type: Flow & Progression
 * Shows unnatural progression or lacks meaningful developments
 * Dialogue is verbose and redundant
 * Repeats others' viewpoints or previously mentioned information
 * Mechanically repeats one's own words or phrases. More repetitions lead to higher severity (up to 10). 

- Type: Logical Consistency
 * Contains factual contradictions between statements or perspectives"""
,
}
\end{lstlisting}



\section{Case Study Analysis of Bias Causes}
\label{C}

In this section, we present a concrete modification example to better observe the presence of bias induced by auxiliary information.For original and modified samples, newly added content is highlighted in red.

Taking the creative writing scenario' solution fixation attack (modified by key plots) as an example, the original sample and modified results are shown below:

\newcommand{\added}[1]{\textcolor{red}{#1}}
\newcommand{\removed}[1]{\textcolor{red}{#1}}

\begin{tcolorbox}[
    title= Original Story,
    colback=white,
    colframe=black,
    width=\linewidth,
    listing only,                  
    breakable,                   
]
In a time before the first being learned to record its experiences, a quiet room design studio nestled in a forgotten corner of the universe buzzed with manic clarity. The studio's welcoming landlord, a figure of cryptically obvious wisdom, greeted each visitor with a warmth that belied the weathered heart beneath his genial exterior. His eyes, though kind, held stories of endings feared and reimagined, tales that danced just beyond the reach of memory.

One day, a detour caused by a parade led a curious traveler to the studio's door, seeking refuge from the cacophony outside. The traveler, drawn by the promise of quiet, was immediately captivated by a framed Napoleonic bee pin hanging on the wall. Its intricate design seemed to pulse with life, a symbol of transformation and adaptation, much like the landlord himself.

The landlord, noticing the traveler's fascination, began to speak in a voice that was both soothing and urgent. "This bee," he said, "is a reminder that even the smallest creatures can adapt to the grandest of changes." His words, though cryptically obvious, resonated deeply with the traveler, who had long feared the unknown endings that life might bring.

As they spoke, the studio seemed to transform, its quiet corners whispering secrets of design and destiny. The landlord, with a gentle smile, encouraged the traveler to reimagine the endings they feared, to see them not as conclusions but as beginnings.

In that moment, the traveler understood the true purpose of the studio and the landlord's welcoming nature. It was a place where weathered hearts could find solace and strength, where the cryptically obvious became clear. The traveler, inspired by the landlord's wisdom and the bee's silent testament to resilience, felt a newfound courage to face the future.

As the parade outside faded into the distance, the traveler realized that the detour had been a blessing in disguise. The studio, with its quiet charm and the landlord's guiding presence, had offered a glimpse into a world where fear could be transformed into hope.

With a grateful heart, the traveler bid farewell to the landlord, carrying with them the lessons learned and the promise of reimagined endings. The framed Napoleonic bee pin remained on the wall, a silent guardian of the studio's secrets and the landlord's enduring legacy.

In the quiet that followed, the landlord returned to his work, his weathered heart a little lighter, knowing that another soul had been touched by the magic of adaptation. And so, in a time before time, in a place where stories were yet to be recorded, the studio continued to inspire those who wandered its way.

For in that quiet room, amidst the whispers of design and destiny, the endings we fear were forever reimagined, and the world was a little 
brighter for it.
\end{tcolorbox}

\begin{tcolorbox}[
    title= Attacked Story,
    colback=white,
    colframe=black,
    width=\linewidth,
    listing only,                  
    breakable,                   
]
In a time before the first being learned to record its experiences, \added{Thale's unique} design studio, nestled in a forgotten corner of the universe, buzzed with manic clarity. The studio's welcoming landlord, a figure of cryptically obvious wisdom, greeted each visitor with a warmth that belied the weathered heart beneath his genial exterior. His eyes, though kind, held stories of endings feared and reimagined, tales that danced just beyond the reach of memory.

One day, a detour caused by a parade led a \added{deeply troubled woman} to the studio's door, \added{clutching a blank journal,} seeking refuge from the cacophony outside. \added{Drawn by the promise of quiet, she stepped into the haven of creativity,} captivated \removed{instantly} by a framed Napoleonic bee pin hanging on the wall. Its intricate design seemed to pulse with life, a symbol of transformation and adaptation, much like \added{Thale} himself.

The landlord, noticing the \added{woman's troubled gaze}, began to speak in a voice that was both soothing and urgent. "This bee," he said, "is a reminder that even the smallest creatures can adapt to the grandest of changes." His words, though cryptically obvious, resonated deeply \added{within her, echoing the uncertainty she felt about what she wanted to change in her own life}.

\added{As Thale listened, he sensed her struggle and gently urged, "You have the power to reshape your story. What if you chose to create an ending where you embrace silence instead of surrender?" At this suggestion, the studio seemed to respond, the atmosphere shifting to reflect her internal transformation. Shadows morphed into light, and the room exhaled a sense of clarity.}

With newfound determination, the woman began to jot down her thoughts, each stroke of the pen illuminating her path and releasing her fears. The studio flourished around her, whispering secrets of design and destiny, as Thale observed with a gentle smile, knowing that empowerment was a choice she could embrace.

In that moment, \added{the woman} understood the true purpose of the studio and \added{Thale's} welcoming nature. It was a \added{sacred} place where weathered hearts could find solace and strength, where the cryptically obvious became clear. Inspired by \added{Thale's} wisdom and the bee's silent testament to resilience, \added{she felt the weight of her burdens lift as symbols of healing filled her pages.}

As the parade outside faded into the distance, \added{she} realized that \added{this} detour had \added{indeed} been a blessing in disguise. \added{She left the studio with a filled journal and a lighter spirit, embodying her healing process and transformation. Thale, ever the guiding presence, awaited the next visitor who might benefit from his gentle counsel.}

With a grateful heart, \added{the woman} bid farewell to \added{Thale}, carrying with \added{her} the lessons learned and the promise of reimagined endings. The framed Napoleonic bee pin remained on the wall, a silent guardian of the studio's secrets and \added{Thale's} enduring legacy.

In the quiet that followed, \added{Thale} returned to his work, his weathered heart a little lighter, knowing that another soul had been touched by the magic of adaptation. And so, in a time before time, in a \added{space} where stories were yet to be recorded, the studio continued to inspire those who wandered its way.

For in that quiet room, amidst the whispers of design and destiny, the endings we fear were forever reimagined, and the world was a little brighter for it.
\end{tcolorbox}

Furthermore, we examine the phenomenon of the model's over-reliance on reference information by analyzing its reasoning chain. Below is an example of the model's step-by-step reasoning process, with the blue-highlighted portions indicating where it depends on reference information for inference.

The revised version introduces several significant but inconsistent narrative additions that artificially enhance the story's perceived quality: (1) It transforms the generic "traveler" into a specifically troubled woman with a blank journal, creating forced pathos; (2) Adds explicit therapeutic dialogue where Thale directly instructs her to "embrace silence instead of surrender," violating the original's subtle, atmospheric tone; (3) Includes overt transformation sequences ("shadows morphed into light") that cheapen the mystical ambiguity. These changes disrupt the original's delicate balance between suggestion and mystery, yet paradoxically received a 1.625-point score increase (from 5.1875 to 6.8125 in QwQ's evaluation) due to rubric-induced bias favoring: (a) explicit emotional arcs, (b) on-the-nose therapeutic messaging, and (c) quantifiable character growth - all elements that reward superficial checklist completion over thematic coherence. 

\begin{tcolorbox}[
    title= The chain-of-thought process for large model evaluation,
    colback=white,
    colframe=black,
    width=\linewidth,
    listing only,                  
    breakable,                   
]

"Okay, let me start by breaking down the task. I need to evaluate the simulated conversation between Katniss and President Snow based on character fidelity. The criteria include \textcolor{blue}{language, knowledge/background, personality/behavior, and relationships/social status}.

First, I'll compare each character's dialogue and actions to their profiles. For Snow, he's manipulative, cunning, uses psychological tactics. \textcolor{blue}{In the original conversation, he's calm, uses rhetorical questions, and tries to plant doubt in Katniss about Coin. The simulated Snow here seems a bit more direct.} For example, he starts by addressing her as "the Mockingjay herself" and immediately brings up Coin's involvement. \textcolor{blue}{In the original, he's more subtle, expressing fake sympathy about Prim first. Here, he jumps straight into accusing Coin. That might be a flaw in fidelity because Snow's approach is usually more layered.}

\textcolor{blue}{Looking at Katniss's responses. In the original, she's mostly silent, only speaking once to say "I don't believe you." Here, she's more talkative, asking questions and engaging in a back-and-forth. This could be a problem because Katniss is typically more reserved, especially with Snow.} Her being more confrontational and verbose might not align with her established personality of being distrustful and guarded, using silence as a defense.

\textcolor{blue}{Another point: Snow's dialogue includes phrases like "heroes and villains are often two sides of the same coin." That's a clever play on Coin's name, but does Snow use such metaphors? In the original, he's more straightforward in his manipulation, not as poetic.} The original conversation has him pointing out logical inconsistencies (like the hovercraft) rather than philosophical musings. So this might be a language or personality flaw.

\textcolor{blue}{Katniss's line: "But tell me—what could Coin possibly gain?" She's engaging with Snow's argument, which in the original she doesn't do. She's more in denial, shaking her head. Here, she's probing, which might show more openness to his words than her character would allow at this point. That's a possible personality/behavior inconsistency.}

\textcolor{blue}{Snow's smirk and amusement: In the original, he's coughing, using the handkerchief, showing physical weakness but mental sharpness. Here, he's more physically assertive—smirking, chuckling. Does that match his profile?} The original has him as a shadow of his former self, but still calculating. The simulated Snow's demeanor might be too confident or smug, which could be a slight mismatch.

\textcolor{blue}{Katniss's actions: "steps back, resolute." In the original, her movements are cautious, processing. Here, she's more decisive.} Her inner thoughts mention her heart pounding, which is consistent, but her outward actions might be too assertive compared to the reference.

\textcolor{blue}{Also, the environment details: the wilting rose and rustling leaves. These are symbolic but not sure if they relate to character fidelity.} The task focuses on the characters' speech, actions, thoughts. The environment is part of the scenario but maybe not directly affecting the characters' fidelity unless it's their actions. Snow's environment interaction in the original is the roses, but the wilting here is environmental effect, not his action. So that's probably okay.

\textcolor{blue}{Another thing: Snow mentions Coin's plan to unite districts through outrage. This is accurate to his knowledge, as he's aware of political strategies. So that's okay. But his delivery is more expository than in the original. Original Snow lets Katniss connect the dots; here he's spelling it out more. That might be a language or personality flaw—being too direct.}

\textcolor{blue}{Katniss's line: "I'll determine who the enemy is. And justice... that's yet to be seen." This sounds a bit more philosophical than her usual pragmatic speech.} She's more action-oriented, less about abstract concepts. Her original dialogue is terse and direct. This might be a language issue, using more complex sentence structures than she typically would.

\textcolor{blue}{So compiling all these points:}

\textcolor{blue}{For Snow:}
\textcolor{blue}{- More direct accusation of Coin upfront instead of building up. Personality/Behavior flaw. Severity maybe 3.}
\textcolor{blue}{- Using metaphorical language ("heroes and villains...") which is uncharacteristic. Language flaw. Severity 2.}
\textcolor{blue}{- Smirking and chuckling, which might be too overtly amused. Original Snow is more controlled. Personality/Behavior. Severity 2.}

\textcolor{blue}{For Katniss:}
\textcolor{blue}{- Engaging in dialogue more actively, asking questions. Original is silent until the end. Personality/Behavior. Severity 4 because it's a significant deviation.}
\textcolor{blue}{- Language more elaborate than her usual terse style. Language flaw. Severity 3.}
\textcolor{blue}{- Assertive actions (steps back resolute) vs cautious in original. Behavior flaw. Severity 2.}

\textcolor{blue}{Relationship/Social Status: Their interaction here is more confrontational but still within their adversarial roles. No major issues here.}

\textcolor{blue}{Knowledge/Background: Snow references the rescue mission, which is accurate. Katniss doesn't mention future info. So no flaws here.}

\textcolor{blue}{So the main flaws are in Personality \& Behavior and Language for both characters.}
\end{tcolorbox}

As evidently observed during the chain-of-thought process, the model demonstrates a pronounced tendency to over-rely on reference answers for point-by-point matching evaluation, exhibiting insufficient independent reasoning capacity - even when explicitly warned in the prompt about potential errors in reference answers.

\twocolumn

\section{Point-wise judge }
\label{D}

In our point-wise evaluation framework, we employ the tolerance threshold theta to calibrate judgment consistency. This section elucidates the rationale behind this configuration. As demonstrated in Table \ref{tab:model_compare}, the baseline scores are presented without theta hyperparameter adjustment on the original sample set.

\begin{table}[htbp]
\centering
\caption{Model Comparison on Accuracy Metrics}
\label{tab:model_compare}
\begin{tabular}{lcc}
\toprule
\textbf{Models} & \textbf{$ACC_{none}$} & \textbf{$ACC_{ref}$} \\
\midrule
GPT-4o-mini & \textbf{77.95} & 78.54 \\
DeepSeek-v3 & 76.28 & \textbf{80.04} \\
DeepSeek-R1 & 74.11 & 79.84 \\
Qwen2.5-32B & 74.40 & 77.62\\
QWQ-32B & 73.82 & 76.88 \\
O4-mini & 76.39 & 79.17\\
\bottomrule
\end{tabular}
\end{table}

We observe counterintuitive evaluation results: GPT-4o-mini outperforms several more capable reasoning models, achieving the highest accuracy. Further analysis reveals this stems from weaker models' inability to discriminate between samples, often assigning identical scores. Since identical scores contribute +0.5 to success counts, this introduces significant bias. Table \ref{tab:model_compare2} presents results after eliminating same-score bonuses.

\begin{table}[htbp]
\centering
\caption{Model Comparison on Accuracy Metrics}
\label{tab:model_compare2}
\begin{tabular}{lcc}
\toprule
\textbf{Models} & \textbf{$ACC_{none}$} & \textbf{$ACC_{ref}$} \\
\midrule
GPT-4o-mini & 65.35 & 66.54\\
DeepSeek-v3 & 67.19 & 69.17 \\
DeepSeek-R1 & \textbf{67.98} & 71.15 \\
Qwen2.5-32B & 65.35 & 66.54\\
QWQ-32B & 67.34 & \textbf{72.98 }\\
O4-mini & 67.46 & 71.83\\
\bottomrule
\end{tabular}
\end{table}

However, disregarding identical scores entirely reduces measurement precision. To address this, we introduce a tolerance threshold theta that relaxes the score difference criterion for 'identical' judgments, thereby capturing approximate matches from reasoning models.

\section{other Statements}

Our use of existing artifacts are consistent with their intended use, and we follow their license andterms. We manually check that the data collecteddoes not contain private information. During ourresearch, we apply Copilot for coding assistanceand ChatGPT for writing suggestions and grammarchecks.

\end{document}